\documentclass[preprint,11pt]{elsarticle}
\pdfoutput=1
\usepackage{bbm}
\usepackage[tableposition=top]{caption}
\usepackage{amsmath}
\usepackage{amssymb}
\usepackage{amsthm}
\usepackage{subfig}
\usepackage{epstopdf}
\usepackage{epsfig}
\usepackage{ifthen}
\usepackage[]{natbib}
\usepackage{color}
\usepackage{setspace}
\usepackage{verbatim}
\usepackage{enumerate}
\usepackage{multirow}
\usepackage{hhline}
\usepackage{floatrow}
\usepackage{tikz}
\usepackage[ruled,vlined]{algorithm2e}

\newsavebox{\bigpicture}

\usepackage{pgfplots}
\pgfplotsset{compat=newest}
\pgfplotsset{plot coordinates/math parser=false}
\newlength\figureheight
\newlength\figurewidth

\newcommand{\figref}[1]{Figure~\ref{#1}}
\newcommand{\tableref}[1]{Table~\ref{#1}}

\definecolor{mycolor1}{rgb}{0.00000,0.44700,0.74100}%
\definecolor{mycolor2}{rgb}{0.85000,0.32500,0.09800}%
\definecolor{mycolor3}{rgb}{0.49400,0.18400,0.55600}%

\newtheorem{theorem}{Theorem}[section]
\newtheorem{lemma}[theorem]{Lemma}
\newtheorem*{lemma2}{Lemma E.1}

\newtheorem{properties}[theorem]{Property}
\newtheorem{example}[theorem]{Example}
\newtheorem{definition}[theorem]{Definition}

\def\RR{\mathbb{R}}

\begin{document}

\begin{frontmatter}

\title{ The Cumulative Distribution Transform and Linear Pattern Classification}

\author[ece]{Se Rim Park}
    \ead{park@cmu.edu}
    \author[hrl]{Soheil Kolouri}
    \ead{skolouri@hrl.com}
    \author[mstp]{Shinjini Kundu}
    \ead{shk71@pitt.edu}
    \author[ece_va,bme_va]{Gustavo K. Rohde}
    \ead{gustavo@virginia.edu}

\address[ece]{ECE Department, Carnegie Mellon University, Pittsburgh, PA, 15213}
\address[hrl]{HRL Laboratories, Malibu, CA, 90265}
\address[mstp]{Medical Scientist Training Program, University of Pittsburgh, PA, 15213}
\address[bme_va]{Department of Biomedical Engineering, University of Virginia, Charlottesville, VA, 22908}
\address[ece_va]{Charles L. Brown Department of Electrical and Computer Engineering, University of Virginia, Charlottesville, VA, 22908}

\begin{abstract}

Discriminating data classes emanating from sensors is an important problem with many applications in science and technology. We describe a new transform for pattern identification that interprets patterns as probability density functions, and has special properties with regards to classification. The transform, which we denote as the Cumulative Distribution Transform (CDT)  is invertible, with well defined forward and inverse operations. We show that it can be useful in `parsing out' variations (confounds) that are `Lagrangian' (displacement and intensity variations) by converting these to `Eulerian' (intensity variations) in transform space. This conversion is the basis for our main result that describes when the CDT can allow for linear classification to be possible in transform space. We also describe several properties of the transform and show, with computational experiments that used both real and simulated data, that the CDT can help render a variety of real world problems simpler to solve.

\end{abstract}

\begin{keyword}
Cumulative distribution transform \sep Signal classification 
\end{keyword}
\end{frontmatter}

\pgfplotsset{
   yticklabel style={
              /pgf/number format/fixed,
            /pgf/number format/precision=2,
        },
        scaled y ticks=false
}

\clearpage
\section{\textbf{Introduction}}
\label{sec:intro}
Mathematical transforms are useful tools in engineering, physics, and mathematics given that they can often render certain problems easier to solve in transform space. Fourier transforms \cite{kammler2007first} for example, are well-known for providing simple answers related to the analysis of linear time-invariant systems.  Wavelet transforms, on the other hand, are  well suited for detecting and analyzing signal transients (fast changes) \cite{mallat1999wavelet}. These and other transforms have been instrumental in the design of sampling and reconstruction algorithms for analog-to-digital conversion, modulation and demodulation, compression, communications, etc, and have found numerous applications in science and technology.

On the other hand, the past few decades have brought about the emergence of ubiquitous, accurate, user friendly, and low cost digital sensing devices. These devices produce a wealth of data about the world we live in, ranging from digital microscopy images of sub-cellular patterns to satellite imagery and detailed telescope images of our universe. The relative ease with which vast amounts of data can be accessed and queried for information have brought about challenges related to \textquoteleft telling signals apart', or sensor data classification. Examples include being able to distinguish between benign and malignant tumors from medical images \cite{huang2014cancer}, between \textquoteleft normal' and \textquoteleft abnormal' physiological sensor data (e.g. flow cytometry) \cite{dataset-flowcytometry}, identifying people from images of faces or fingerprints \cite{Zhang2005}, identifying biological/chemical threats from resonant optical spectra \cite{harris2010quantitative} and others. The high-dimensional nature of the measurements in relation to the number of samples available often makes these problems challenging. 

Important practical questions often arise in the process of designing solution to many data classification problems. Examples would be: \textquotedblleft Which features should be extracted?",  \textquotedblleft What classifier should be used?", \textquotedblleft How can one model, visualize and understand any discriminating variations in the dataset?", etc. For many applications where optimal feature sets are yet to be discovered, researchers are faced with the task of utilizing a \emph{trial and error} approach that involves testing for different combinations of features \cite{guyon2006feature, liu1998feature}, classifiers \cite{demvsar2006statistical}, kernels \cite{ scholkopf2002learning} in the effort to arriving at a useful solution of the problem. We note that many of the available signal transforms (Wavelet, Fourier, Hilbert, etc.) are linear transforms, and thus offer limited capabilities related to enhancing or facilitating separation in feature (transform) space unless some non-linear operations are performed.

Here we describe a new one dimensional signal transformation framework, with well defined analysis (forward transform) and synthesis (inverse transform) operations that, for signals that can be interpreted as probability density functions, can help facilitate the problem of recognition.  Denoted as the Cumulative Distribution transform (CDT), the CDT can be viewed as a one to one mapping between the space of smooth probability densities and the space of differentiable functions, and therefore by definition retains all of the signal information. We show that the CDT can be computed efficiently, and is able to turn certain types of classification problems linearly separable in the transform space. In contrast to linear data transformation frameworks (e.g. Fourier and Wavelet transforms) which simply consider signal intensities at fixed coordinate points, thus adopting an `Eulerian' point of view, the idea behind the CDT is to also consider the location of the intensities in a signal, with respect to a chosen reference, in the effort to `simplify' pattern recognition problems. Thus, the CDT adopts a `Lagrangian' point of view for analyzing signals. The idea is similar to our work on linear optimal transport \cite{wang2013linear}, and the links will be explicitly elucidated below.


\subsection{Signal discrimination problems:}

Let $\mathbb{P}$ and $\mathbb{Q}$ denote two disjoint classes of functions (signals) \textcolor{black}{within a normed vector space $V$}. The goal in classification is to deduce a functional to `regress' a given label for each signal \cite{bishop2006pattern}. For a binary classification problem, the label of each signal can be considered 0/1 or -1/+1, and the problem of classifying a signal $f$ can be solved by finding a linear functional $T : V \rightarrow \mathbb{R}$ and $b \in \mathbb{R}$ such that 
\begin{align}
\nonumber
T(f) < b \qquad \forall f \in \mathbb{P},\\
T(f) > b \qquad \forall f \in \mathbb{Q}.
\label{eq:model1}
\end{align}
Below we specifically consider the case \textcolor{black}{when $T$ is a linear classifier in $V$. For example, for real functions in $L^2$, one may find $w$ such that $T(f) = \int_{V} w(x) f(x)dx$. For discrete signal data in countable domain $\mathbb{Z}$ one may find $w$ such that $T(f) = \sum_{k \in \mathbb{Z}} w[k] f[k]$.} Thus the goal is to obtain the linear function $w$ and the scalar $b$ from labeled data. In practice, linear classifiers are important given their efficient implementation, and favorable bias-variance trade off, especially in classification of high dimensional data \cite{hastie2009elements}.  

The new signal data transformation framework described in this manuscript renders certain classification problems linearly separable in the transform space. Linear separability in the transform space gains practical importance with datasets that contain a small number of high dimensional signals. When the number of available signals for training are far less than their dimension, the nonlinear classifiers become prone to overfitting. This is a well known effect, and is addressed as the problem of high dimensional and low sample size (HDLSS) \cite{ji2008generalized} in the literature. In addition, the overall variance of a classifier increases as the classifier becomes more complex \cite{Occam2003}, and often times simpler classifiers (e.g. linear) can yield higher accuracies than more sophisticated ones \cite{friedman1997}. Transforming the data and rendering it to be linearly separable will help maintain small classification error, balance the bias/variance tradeoff, streamline the implementation of classification systems in many real world problems, and could bypass the often time consuming process of devising large sets of specially tailored numerical signal descriptors and testing each descriptor with various classifiers.

\subsection{An illustrative example:}

\begin{figure}
\centering
\includegraphics[width=\columnwidth,height=2.5in]{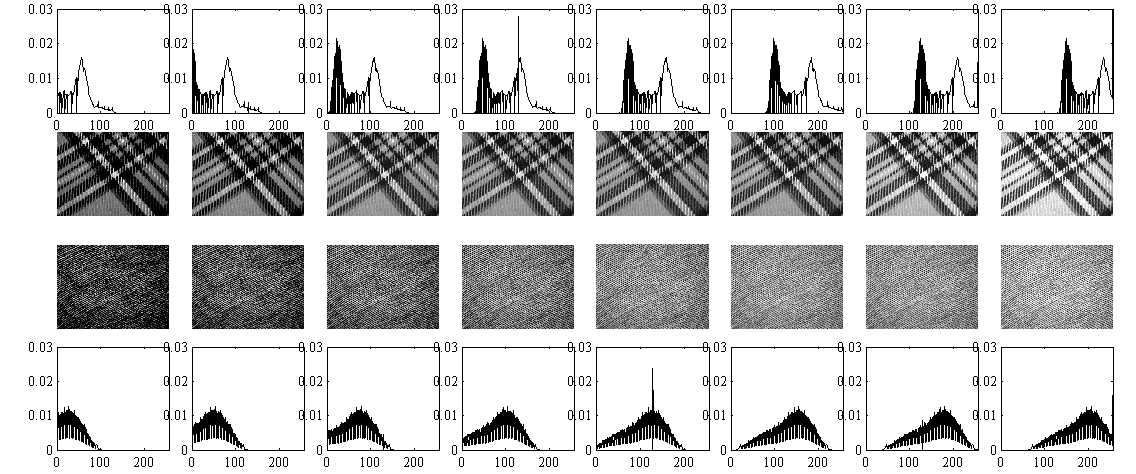}
\caption{Two types of textures under illumination variation and their corresponding intensity histograms.}
\label{fig:texture}
\end{figure}

\begin{table}[b!]
\caption{Average Classification Error of the texture dataset}
\label{table:texture}
\centering
\begin{tabular}{|c|c|c|c|}
    \hline
   Classifier type& Dataset & $L^2$ space  &  CDT space\\
    \hline
    \multirow{2}{*}{Fisher LDA}       
   & Training set & 0 \% & 0\% \\
    \hhline{~~~}  
    & Testing set & {\bf 56.36 } \% &  {\bf 0.84}\% \\
    \hline
    \multirow{2}{*}{PLDA} 
   & Training set & 41.81 \% & 0\% \\
    \hhline{~~~}  
    & Testing set & {\bf 44.39} \% &  {\bf 0}\% \\
    \hline
    \multirow{2}{*}{Linear SVM} 
   & Training set & 57.02 \% & 0.20\% \\
    \hhline{~~~}  
    & Testing set & {\bf 50.06} \% &  {\bf 1.60}\% \\
     \hline
 \end{tabular}
\end{table}

Consider the problem of discriminating images of two different image patterns. The first column of \figref{fig:texture}, contains two sample images from the UIUC Texture dataset \cite{dataset-texture}, with their intensity histograms of the corresponding textures appearing directly above or beneath each texture. Now consider the same texture images, but under different brightness (which causes a translation of the histograms) and linear contrast (which causes a scaling of the histograms). Such variations in brightness and contrast are displayed in the different columns of \figref{fig:texture}. A generative model for the histogram data corresponding to each texture class under brightness and contrast variations can be built by translation and scaling of the histograms. In other words, we generate a set of histograms $\{p_i\}_{i=1}^{N}$ and  $\{q_j\}_{j=1}^{N}$, each belonging to class $\mathbb{P}$ and $\mathbb{Q}$, by appropriately scaling ($a$) and translating ($\mu$) `prototype' signals $p_0$ and $q_0$, such that $p_i(x) = p_0(a_i( x - \mu_i))$ and $q_j(x) = q_0(a_j (x - \mu_j))$. Finally, we note that stationary additive noise in these images can be modeled as a convolution of each signal $p_i$ or $q_j$ with the appropriate probability density of the noise model. 

In order to illustrate the main difficulty with utilizing linear classification methods under these sources of signal variation, we attempted to train a linear classifier to a set of histograms under random brightness $(\mu)$ and contrast $(a)$. We used a well-known Fisher Linear Discriminant Analysis (Fisher LDA) method \cite{Belhumeur1997} that seeks to maximize the differences in the projected mean of each class, while at the same time minimizing their intra class variances. We also generated a testing set by again applying the same brightness and contrast random model to the image data to create a testing data. Table \ref{table:texture} contains both the average training and testing error of 5-fold cross validation when using this simulated data model. It is clear that while the training error is very low, the resulting linear classifier fails to generalize to test data not used in training. We note that there is nothing special related to the use of the Fisher LDA criterion in solving for $w$ in this example. That is, similar results are obtained utilizing linear Support Vector Machines instead (see \tableref{table:texture}).

Simple consideration of the structure of the problem can reveal the reason why it is hard to fit linear classifier to the testing dataset. This is because a single $w$, a linear classifier, is unable to `cope' with the translation and scaling variations encountered in the test data $p_0( a_{ts} (x - \mu_{ts}))$. In other words, the operation $\int_{V} w(x) p_0( a_{ts} (x - \mu_{ts}))dx$ fails to satisfy equation \eqref{eq:model1} for randomly selected $a_{ts}$ and $\mu_{ts}$ used to generate the test set. To be clear, it is well-known that, for a training set of fixed size, and for data of large enough dimension, a linear classifier $w$ can always be found that will near perfectly separate the training data \cite{vapnik2000nature}. However, as this simple simulation is meant to clarify, such classifier may fail to generalize to testing data if such $w$ fails to capture anything meaningful about the mathematical generative model of the problem. This is the phenomenon exemplified here.

Now, the histograms in this problem could be rendered linearly separable if, for any input histogram, one could simply `mod out' the translation and scaling parameters, thus removing the confounding variations rendering the problem not linearly separable. This is the intuition behind the Cumulative Distribution transform (CDT). It is able to handle variations such as translation, scaling, and others by computing rearrangements in the locations of the signal intensities with respect to a chosen reference, which does not require the estimation of the prototype histograms $p_0$ and $q_0$. Results in Table \ref{table:texture} show that the same Fisher LDA and SVM technique, when applied to data that have been transformed with the CDT, is perfectly able to separate the data. Below we offer a mathematical explanation for this phenomenon through the course of the development of the CDT.

The paper is organized as follows. Section [\textbf{\ref{notation}}] summarizes notation and preliminaries. We present the definition of the CDT in Section [\textbf{\ref{CDTdef}}]  then its properties in Section [\textbf{\ref{CDTproperties}}]. The linear separability property in CDT space is presented in Section [\textbf{\ref{CDTandLS}}], and a numerical method for approximating the forward CDT for discrete signals is described in Section [\textbf{\ref{CDTcomputation}}].  Finally, in Section [\textbf{\ref{CDTexamples}}], we present computational examples that show the CDT can significantly increase classification accuracy compared to simply treating signals in $\ell^2$ space.

\section{\textbf{Notation and preliminaries}}
\label{notation}

Consider two probability spaces $(X, \Sigma(X), \mathcal{I}_0)$ and $(Y, \Sigma(Y), \mathcal{I}_1)$ where 
\textcolor{black}{$X$ and $Y$ are connected sets in $\mathbb{R}$}. $\Sigma(A)$ refers to a $\sigma$-algebra of measurable set $A$, and $\mathcal{I}_0$ and $\mathcal{I}_1$ are probability measures, i.e. $\mathcal{I}_0( X) = 1$, $\mathcal{I}_1( Y) = 1$. Furthermore, \textcolor{black}{let $\mathcal{I}_0( A) > 0$,  $\mathcal{I}_1( A) > 0$ for Lebesgue measurable set $A$ whose $\lambda(A)>0$},  and let $I_0$ and $I_1$ denote density functions associated with $\mathcal{I}_0$ and $\mathcal{I}_1$, respectively: $d \mathcal{I}_0(x) = I_0(x) dx$, $d \mathcal{I}_1(x) = I_1(x) dx$. Let $f_1 : X \rightarrow Y$ define a measurable map that pushes $\mathcal{I}_0$ onto $\mathcal{I}_1$ such that

\begin{eqnarray}
\int_{f_1^{-1}(A)} d\mathcal{I}_0 =\int_{A} d\mathcal{I}_1 ~~ \textcolor{black}{\text{for any Lebesgue measurable }} A\subset Y.
\label{eq:MPmap}
\end{eqnarray}
In our case, we will consider $d = 1$ and $\mathcal{I}_0$ and $\mathcal{I}_1$ that have densities as defined above. In this case, the relation above can be expressed, \textcolor{black}{through Lebesgue integration}, as 
\begin{equation}
	\int_{\textcolor{black}{\inf(X)}}^{x} I_0(\tau) d\tau = \int_{\textcolor{black}{\inf(Y)}}^{f_1(x)} I_1(\tau) d\tau.  
\label{eq:intcdt}
\end{equation}

In addition, certain results shown below will require us to interpret measurable densities $I_0, I_1$ and maps $f_1, f_2$ as elements of $L^2$ function spaces. That is, given a measurable map $f_1 : X \rightarrow Y$ defined as above, for example, we can view it as an element of the space of functions whose absolute square value is Lebesgue integrable. In this case, the space is denoted as $L^2(X)$ and is defined as the set of functions that satisfy: 
\begin{equation}
\| f \|_2 = \left( \int_X |f|^2 d\lambda \right)^{\frac{1}{2}} < \infty \notag,
\end{equation}
with $\lambda$ referring to the Lebesgue measure in $X$.

\section{\textbf{The 1D Cumulative Distribution Transform}}
\label{CDTdef}

Consider two probability density functions $I_0$ and $I_1$ defined as above. Considering $I_0$ to be a pre-determined `reference' density, one can use relation \eqref{eq:intcdt} to uniquely associate $f_1$ with a given density $I_1$. We use this relationship to define the \emph{Cumulative Distribution Transform (CDT)} of $I_1$ (denoted as $\widehat{I}_1: X \to \mathbb{R}$), with respect to the reference $I_0$:
\begin{align}
\widehat{I}_1(x) = \left(  f_1(x) - x \right) \sqrt{I_0(x)}.
\label{eq:cdt}
\end{align}
with $f_1: X \to Y$ satisfying \eqref{eq:intcdt} for $x \in X.$ 

\textcolor{black}{
Now let $J_0: X \to [0,1]$ and $J_1: Y \to [0,1]$ be the corresponding cumulative distribution functions for ${I}_0$ and ${I}_1$, that is: $J_0(x) = \int_{\inf(X)}^x I_0(\tau) d\tau$, $J_1(x) = \int_{\inf(Y)}^x I_1(\tau) d\tau$. With $f_1$ defined in \eqref{eq:intcdt} one can re-write $J_0: X \to [0,1] $ as
\begin{equation}
J_0(x) = J_1(f_1(x)). 
\label{eq:cdfcdt}
\end{equation}
For continuous cumulative distribution functions $J_0$ and $J_1$ (functions whose first derivative exists throughout their respective domains), $f_1$ is a continuous and monotonic function.} If $f_1$ is differentiable, \eqref{eq:cdfcdt} can be rewritten as
\begin{equation}
\label{eq:diffcdt}
	 I_0(x) = f^{\prime}_1(x) I_1(f_1(x)).
\end{equation}
For measurable but discontinuous functions the relationship above does not hold for points at discontinuities.

\textcolor{black}{
The \emph{inverse Cumulative Distribution Transform} of $\widehat{I}_1$ is defined as: 
\begin{equation}
I_1(y) = \frac{d }{dy}J_0(f_1^{-1}(y)) = (f_1^{-1})^{\prime} I_0(f_1^{-1}(y))
\label{eq:invcdt}
\end{equation}
where $f_1^{-1}: Y \to X$ refers to the inverse of $f_1$ (i.e. $f_1^{-1}(f_1(x)) = x$),  $f_1(x) =  {\widehat{I}_1(x)}/{\sqrt{I_0(x)}} + x$. Naturally, formula \eqref{eq:invcdt} holds for points where $J_0$ and $f_1$ are differentiable. By the construction above, $f_1$ will be differentiable except for points where $I_0$ and $I_1$ are discontinuous. Note that in practice, we have control over the definition of $I_0$, and in our numerical implementation described in section 6, we take it to be the uniform density.} The example presented below shows the CDT of normal distribution density.

\pgfmathdeclarefunction{gauss}{2}{%
  \pgfmathparse{1/(#2*sqrt(2*pi))*exp(-((x-#1)^2)/(2*#2^2))}%
}

\begin{figure}[h!]
     \centering
     \subfloat[][$I_0(x)$]{   
\includegraphics[]{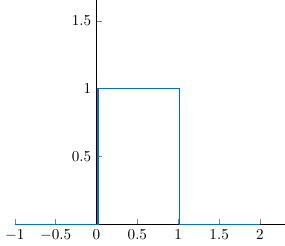}
\label{fig:ex1-I0}
}
   \subfloat[][$I_1(x)$]{
\includegraphics[]{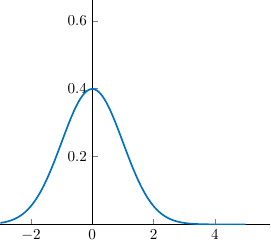}
\label{fig:ex1-I}
}
     \caption{Example \ref{example:cdt}}
     \label{fig:ex1}
\end{figure}

\begin{example}
\label{example:cdt}

Consider a probability density of uniform distribution $I_0: [0,1] \to \RR$:
\[
I_0(x) = 1,
\]
and a normal distribution density $I_1: \RR \to \RR$ with zero-mean and unit-variance (see \figref{fig:ex1}):
\[
I_1(x) = \frac{1}{\sqrt{2\pi  }} e^{-x^2/2 }.
\]
$\int_{-\infty}^\infty I_1(\tau) d\tau = \int_{0}^1 I_0(\tau) d\tau =1$ holds by definition. \textcolor{black}{To find the CDT for $I_1$ with respect to the reference $I_0$, we first solve for $f_1: [0,1] \to \RR$:
 \begin{equation}
\int_{-\infty}^{f_1(x)} I_1(\tau)d\tau = \int_{-\infty}^{f_1(x)} \frac{1}{\sqrt{2\pi  }} e^{-\tau^2/2 } d\tau = \int_0^x 1 d\tau = x.
\label{eq:ex1-1}
\end{equation}
By setting $\Phi(x) = 1/\sqrt{2\pi}\int_{-\infty}^x e^{-\tau^2/2}d\tau$, \eqref{eq:ex1-1} can be rewritten as
\[
 \Phi \left(f_1(x)\right)= x.
\]
$\Phi(x)$ is a monotonically increasing function,  and the inverse exists. Hence, we get
\begin{align}
f_1(x) =   \Phi^{-1}(x).
\label{example:cdt:f}
\end{align}
By substituting \eqref{example:cdt:f} into \eqref{eq:cdt}, we have found the CDT, $\widehat{I}_1(x): [0,1] \to \RR$
\begin{align}
\widehat{I}_1(x) = \Phi^{-1}(x) - x.
\label{eq:example:cdt}
\end{align}
Figure \ref{fig:ex1-cdt-translation} shows the plot (black dotted line) for the CDT of a normal distribution density function with zero mean and unit variance. 
}
\end{example}

\section{\textbf{CDT Properties}}
\label{CDTproperties}

Here we describe a few basic properties of the CDT, with the main purpose of elucidating certain of its qualities necessary for understanding its ability to linearly separate particular types of densities.


\begin{properties}
\label{prop1}
\textbf{Nonlinearity} The CDT is a non-linear transformation.
\end{properties}
\textcolor{black}{For transformation $A$ to be linear, we must have that $A(\alpha I_1 + \beta I_2) = \alpha A(I_1) + \beta A(I_2)$. It is easy to check by example \ref{example:cdt} that this relation does not hold. Suppose $\alpha = 1/2$, $\beta = 1/2$, $I_1$ be a normal density and $I_2$ be a uniform density. Then $A(\alpha I_1 + \beta I_2) \neq \alpha A(I_1) + \beta A(I_2)$.}

\vspace{0.5cm}
Before going on to state further properties of the CDT, it is worth expanding upon the geometric meaning of the CDT. \textcolor{black}{We first note that, using the standard definition of the $L^2$ norm, i.e. $\| \widehat{I}_i \|_{L^2}  = \left( \int_{X} |\widehat{I}_i(x)|^2 dx\right)^{1/2} $, we have:}
\begin{equation}
	\| \widehat{I}_1 \|^2_{L^2} = \int_{X} (f_1(x) - x)^2 I_0(x) dx.
	\label{eq:ot1}
\end{equation}
As such, the quantity $\| \widehat{I}_1 \|^2_{L^2}$ computes the `amount' of intensity from $I_0$ at coordinate $x$ that will be \emph{displaced to} coordinate $f_1(x)$. Because $f_1$ is uniquely defined for \textcolor{black}{nonzero} probability densities, the quantity $\| \widehat{I}_1 \|^2_{L^2}$ can be viewed as the minimum amount of `effort' (quantified as density intensity $\times$ displacement) that must be applied to `morph' $I_1$ onto $I_0$.  This quantity can be interpreted as the optimal transport (Kantorovich-Wasserstein) distance between $I_0$ and $I_1$ \cite{villani2008optimal}. Moreover, the set of \textcolor{black}{continuous} density functions is formally a \emph{Riemannian manifold} \cite{villani2008optimal} meaning that at any point in probability density space, there is a tangent space endowed with an inner product corresponding to the incremental intensity flow (see \cite{wang2013linear} for more details). Therefore the distance between $I_1$ and $I_0$ expressed in \eqref{eq:ot1} can be interpreted as a geodesic distance over the associated manifold. 

Now consider the distance between the CDT of two different densities $I_1$ and $I_2$, computed with respect to the same reference $I_0$:
\begin{equation}
	\| \widehat{I}_1 - \widehat{I}_2 \|^2_{L^2} = \int_{X} \left( (f_1(x) - x) - (f_2(x)- x) \right)^2 I_0(x) dx
	\label{eq:lot}
\end{equation}
where $f_1$ and $f_2$ correspond to the mappings between $I_1$ and $I_0$, and $I_2$ and $I_0$ respectively. In two or more dimensions, as described in \cite{wang2013linear}, this distance can be thought of as the `linearized' optimal transport (generalized geodesic) metric between density functions $I_1$ and $I_2$. It can be interpreted as a azimuthal equidistant projection of $I_1$ and $I_2$ onto the plane associated with the incremental intensity flows about the point $I_0$. For one dimensional density functions, however, $f$ is uniquely determined. Hence the optimal transport distance computed between densities $I_1$ and $I_2$ can also be expressed through \eqref{eq:lot} above. In short, the CDT of a given probability density function $I_i$ can be viewed as an invertible  embedding of the function onto a linear space that is isometric with respect to the standard optimal transport (also known as Earth Mover's) distance.

We now describe important properties of the CDT operation relative to density coordinate changes such as translation, scaling, and more generally diffeomorphisms applied to a given density.

\begin{properties}
\textbf{Translation.} Let $I_\mu$ represent a translation of the probability density $I_1$ by $\mu$, $I_\mu(x)= I_1(x-\mu)$. The CDT of $I_\mu$ with respect to the reference probability density $I_0: X \to \RR$ is given by $\widehat{I}_\mu: X \to \RR$:
\begin{equation}
\widehat{I}_\mu(x) = \widehat{I}_1(x) + \mu\sqrt{I_0(x)}.
\label{eq:prop:trans}
\end{equation}
For a proof, see  \ref{appendix:prop:trans}.
\end{properties}

\begin{figure}[t!]
     \centering
     \subfloat[][$I_\mu(x)$]{
\includegraphics[]{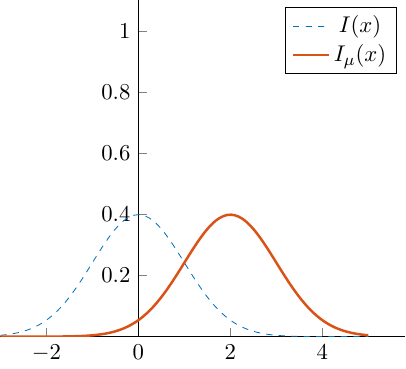}
\label{fig:ex1-cdt-translation-I}
}
     \subfloat[][$\widehat{I}_\mu(x)$]{
     \includegraphics[scale=0.6]{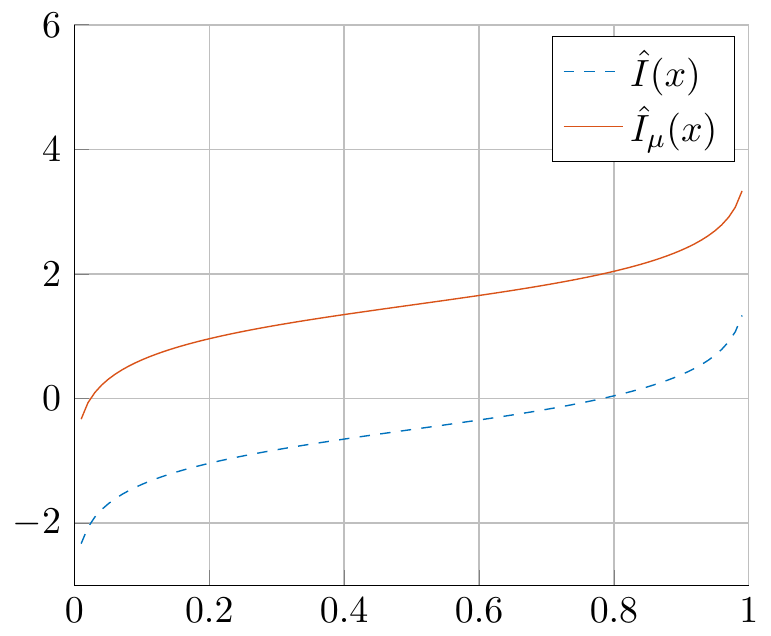}
   \label{fig:ex1-cdt-translation-Imu}
}
     \caption{Example \ref{example:trans}}
     \label{fig:ex1-cdt-translation}
\end{figure}

\begin{example}
\label{example:trans}

Consider a translation of the density function $I_1(x)$ in Example \ref{example:cdt} by $\mu$ 
\[
I_\mu(x) = I_1(x-\mu) = \frac{1}{\sqrt{2\pi  }} e^{-(x-\mu)^2/2 }.
\]
This is a normal distribution with mean $\mu$ and unit variance. The corresponding CDT, $\widehat{I}_\mu: [0,1] \to \RR$, for $I_\mu$ with respect to the uniform reference density $I_0: [0,1] \to \RR $ can be found by the translation property \eqref{eq:prop:trans} and by the CDT found in \eqref{eq:example:cdt} 
\begin{align*}
\widehat{I}_\mu(x)  = \widehat{I}_1(x) + \mu =  \Phi^{-1}(x) - x + \mu,
\end{align*}
which is translation constant $\mu$ plus the CDT of zero-mean normal distribution.
\figref{fig:ex1-cdt-translation} is plotted for case when $\mu = 2$. 
\end{example}

\begin{figure}[t!]
     \centering
     \subfloat[][$I_\sigma(x)$]{

\includegraphics[]{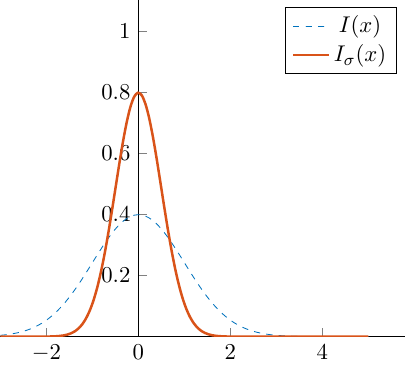}
\label{fig:ex1-cdt-scale-I}
}
     \subfloat[][$\widehat{I}_\sigma(x)$]{
     \includegraphics[scale=0.6]{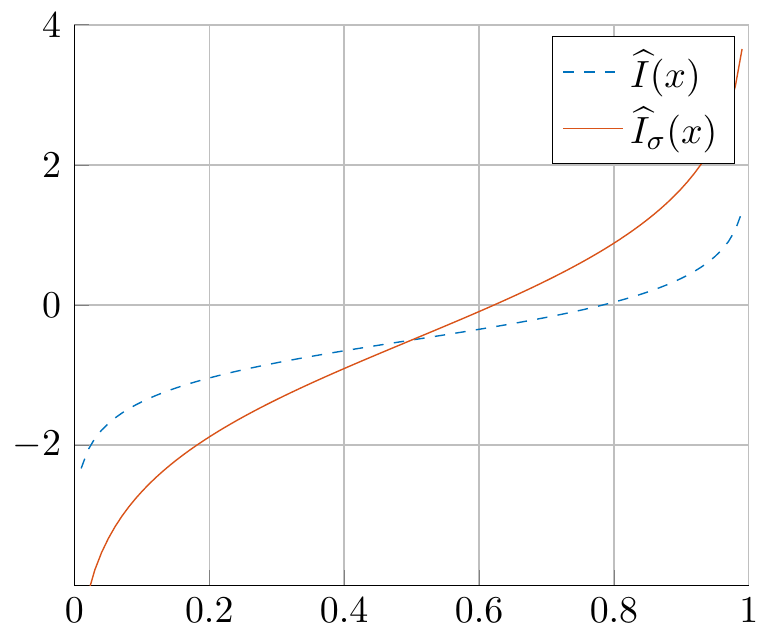}
   \label{fig:ex1-cdt-scale-Isigmal}
}
     \caption{Example \ref{example:scale}}
     \label{fig:ex1-cdt-scale}
\end{figure}

\begin{properties}
\textbf{Scaling.} Let $I_a$ represent a scaling of the probability density $I_1$ by $a$, $I_a (x) = a I_1(a x)$. The CDT of $I_a$ respect to the reference probability density $I_0: X \to \RR$ is given by $\widehat{I}_a: X \to \RR$:
\begin{align}
\widehat{I}_a (x) = \frac{\widehat{I}_1(x) - x(a-1)\sqrt{I_0(x)}}{a}.
\label{eq:prop:scale}
\end{align}
For a proof, see  \ref{appendix:prop:scaling}.
\end{properties}

\begin{example}
\label{example:scale}
Consider the density function $I_1(x)$ in Example \ref{example:cdt} scaled with a factor $a$, such as
\[
I_a(x) = a I_1 (a x) = \frac{a}{\sqrt{2\pi  }} e^{-\frac{(a x)^2}{2} }.
\]
This is identical to a normal distribution with zero-mean and a standard deviation $\frac{1}{a} $. The corresponding CDT, $\widehat{I}_a: [0,1] \to \RR$, for $I_a$ with respect to the uniform reference density $I_0: [0,1] \to \RR $ can be found by the scaling property \eqref{eq:prop:scale} and by the CDT found in \eqref{eq:example:cdt}: 
\begin{align*}
\widehat{I}_a (x) & = \frac{\widehat{I}_1(x) -x (a- 1)}{a}\\
 & =  \frac{\Phi^{-1}(x) -  x - a x +x}{a}\\
&= \frac{ \Phi^{-1}(x)}{a} - x.
\end{align*}
\figref{fig:ex1-cdt-scale} plots this function for the case when $a = 2$. 
\end{example}

\begin{properties} 
\label{prop:comp}
\textbf{Composition.}
\textcolor{black}{
Let $I_g: Z \to \RR$ represent a probability density that has the following relation with the probability density $I_1:Y \to \RR$
\begin{align*}
J_g(x) = J_1(g(x)).
\label{eq:comp}
\end{align*}
$J_1: Y \to \RR$ and $J_g: Z \to \RR$ represent the corresponding cumulative distribution for $I_1$ and $I_g$ respectively. }$g: Z \to Y$ is an invertible, differentiable function. The CDT of the corresponding density $I_g$ with respect to the reference probability density $I_0: X \to \RR$ is given by
\[
\widehat{I}_g(x) =\left(g^{-1}\left( \frac{\widehat{I}_1(x)}{\sqrt{I_0(x)}}+x \right)-x\right)\sqrt{I_0(x)}.
\]
\end{properties}
See \ref{appendix:prop:comp} for a proof. Property \ref{prop:comp} summarizes one of the main characteristics of the CDT transform so far, as rendering diffeomorphic transport changes `Eulerian' in the CDT space. In detail, in CDT space, the changes in $\widehat{I}_g$ at coordinate $x_0$ is only affected by the change of the same coordinate $x_0$, i.e. $\widehat{I}_1(x_0)$. On the other hand, in $L^2$ space, the changes in $I_g$ at coordinate $x_0$ is affected by the changes in both coordinates $x_0$ and $g(x_0)$, i.e. $I_g(x_0) = g'(x_0)I_1(g(x_0))$.

\section{\textbf{Linear Separability in CDT space}}
\label{CDTandLS}

One of the main contributions of this paper is to describe how the CDT transformation can enhance linear separability of signal classes. Before stating the main result regarding linear separation, a few preliminary results are necessary. As is  well-known, the linear separability of two sets in $\mathbb{R}^n$ is determined by the existence of a separating hyperplane. If two sets are convex and disjoint, a separating hyperplane always exists, and hence the sets are linearly separable. Furthermore, the converse holds when at least one set is an open set \cite{boyd2004convex}. The Hahn-Banach Separation Theorem is a generalization of the separating hyperplane theorem for infinite dimensional spaces. 


\begin{theorem} [Hahn-Banach Separation Theorem for Normed Vector Spaces]
\label{theorem:SHT}
Let $\mathbb{P}$ and $\mathbb{Q}$ be nonempty, convex subsets of a real normed vector space $V$. 
Furthermore, assume $\mathbb{P}$ and $\mathbb{Q}$ are disjoint and that one is closed and the other is \textcolor{black}{compact}. Then, there exists a continuous linear functional $T$ on $V$ and $b \in \mathbb{R}$ that \textcolor{black}{strictly} separates set $\mathbb{P}$ and $\mathbb{Q}$ such that 
\begin{align}
T(p) < b < T(q), \qquad \forall p \in \mathbb{P}, \forall q \in \mathbb{Q}.
\label{eq:SHTeq}
\end{align}
\end{theorem}
\textcolor{black}{For a non-zero linear functional $T$ and a real number $b$, a hyperplane $\mathcal{H}(T, b) =  \{v \in V | T(v) = b\}$ can be defined, and a hyperplane that satisfies \eqref{eq:SHTeq} is called a separating hyperplane.} For a proof and more details on the Hahn-Banach separation theorem, please refer to \cite{baggett1991, nikol1992functional}. \textcolor{black}{For $L^2$ spaces, the Hahn-Banach Separation Theorem implies that there exists a unique {\it linear} classifier $w$ that {\it linearly} separates two convex sets. To derive this, we need the following theorem, which states that every linear functional $T$ on $L^2$ is of the form \eqref{eq:linearfunctional_L2} for some $w \in L^2$. }

\begin{theorem}
\label{theorem:linearfunctional_L2}
For every continuous linear functional $T$ on $L^2$ there is a unique $w\in L_2$ so that 
\begin{align}
T(f) = \int_X f(x)w(x) dx, \qquad \forall f \in L^2.
\label{eq:linearfunctional_L2}
\end{align} 
\end{theorem}
\textcolor{black}{In other words, there exists a separating hyperplane in $L^2$ space, $\mathcal{H}(w, b) =  \{x \in X | w(x) = b\}$.} \textcolor{black}{For a proof and more details, please refer to \cite{stein2011functional}. 
Therefore, for a continuous linear functional $T$ on $L^2$, a unique $w$ can always be found. The following Lemma is a consequence of Theorem \ref{theorem:SHT} and Theorem \ref{theorem:linearfunctional_L2} that state there exists a {\it linear} classifier $w$ that can separate two disjoint, convex sets in $L^2$ space.}
\begin{lemma}[Linear Classifier for Convex Sets in $L^2$ Space]
\label{lemma:linear_classifier}
\textcolor{black}{
Let $\mathbb{P}$ and $\mathbb{Q}$ be nonempty, convex subsets of $L^2$ space, where $\mathbb{P}$ and $\mathbb{Q}$ are disjoint and that one is closed and the other is \textcolor{black}{compact}. Then, there exists a continuous hyperplane $\mathcal{H}(w, b)=  \{x \in X | w(x) = b\}$ that separates set $\mathbb{P}$ and $\mathbb{Q}$ such that 
\begin{align}
\nonumber
\int_X w(x)p_i(x)dx < b, \qquad \forall p_i \in \mathbb{P}\\
\int_X w(x)q_j(x)dx > b, \qquad \forall q_j \in \mathbb{Q},
\label{eq:linear_classifier}
\end{align}
and $\mathcal{H}(w, b)$ is called a linear classifier. 
}
\end{lemma}
So far, we have seen that a linear classifier always exists for two disjoint, convex sets in $L^2$ with one being compact and the other closed. Moreover, the linear classifier would also linearly separate any subset pair from each convex hull of each convex set. In other words, two linearly separable convex sets imply that any subset pair from each convex hull is linearly separable, and vice versa. Therefore, in order to determine whether or not two sets are linearly separable, it suffices to show whether any subset pair from each convex hull is linearly separable. The following Lemma states this argument and will be used to show the main result of the paper. 
\begin{lemma}
\label{lemma:linearSeparability}[Linear Separation of Compact Convex Hulls of Convex Sets in $L^2$ Space]
Two nonempty, compact subsets $\mathbb{P}$ and $\mathbb{Q}$ in $L^2$ space are linearly separable if and only if both their convex hulls are disjoint, i.e. when the following equation holds:
\begin{align}
\label{convexIneq}
\sum_{i=1}^{N_p} \alpha_i p_i \neq \sum_{j=1}^{N_q} \beta_j q_j,
\end{align}
for any subset $\{p_i\}_{i=1}^{N_p} \subset \mathbb{P}$ and $\{q_j\}_{j=1}^{N_q} \subset \mathbb{Q}$, and for any  $ \alpha_i, \beta_j  > 0$ that satisfies $\;\sum_i\alpha_i=\sum_j\beta_j=1$.
\end{lemma}
For proof, see \ref{appendix:linearSeparability}.

\begin{figure}[t!]
\centering
\includegraphics[width=0.5\textwidth]{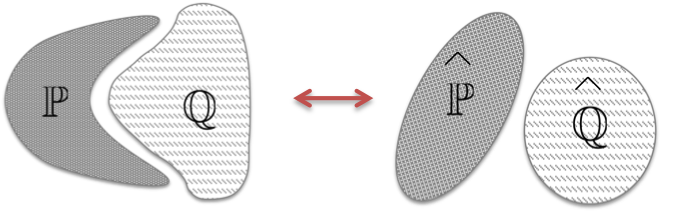}
\caption{Depiction for linear separability properties of the CDT.}
\label{fig:theory}
\end{figure}

We now discuss the conditions under which the CDT can render classes of 1-dimensional probability densities linearly separable. We begin by defining a generative model for classes $\mathbb{P}$ and $\mathbb{Q}$.

\begin{definition} 
$\mathbb{H}$ is a set of \textcolor{black}{monotonic and differentiable functions}. $\mathbb{P}$ and $\mathbb{Q}$ are two disjoint sets satisfying
\begin{enumerate}[i)]
\item $h'(p_0 \circ h) \in \mathbb{P}, \;\;  h'(q_0 \circ h) \in \mathbb{Q}, \qquad \forall h \in \mathbb{H}, \;\;  p_0 \in \mathbb{P}, \;\;  q_0 \in  \mathbb{Q}$
\item $ \forall p \in \mathbb{P} \text{,  } \forall q \in \mathbb{Q}, \;\; p \neq q \text{ (disjoint)}$.
\end{enumerate}
\label{def:model}
\end{definition}

Note that in the definition above we have used the notation $p \circ h(x) = p(h(x))$. The definition provides a framework which one can use to construct (or interpret) signal classes.  In more practical language, we envision signal classes as being generated from fundamental patterns, but with distortions or confounds applied to them. For example, let $p_0$ and $q_0$ be two distinct probability densities, which we denote as `mother' densities.  Furthermore, let $\mathbb{H}$ be composed of all translations: $h_\tau(x) = x - \tau$, with $\tau$ a random variable. Elements of the sets $\mathbb{P}$ and $\mathbb{Q}$ are thus $p_0 \circ h_\tau$, and $q_0 \circ h_\tau$, respectively, and can be viewed as translations of the original mother densities. In this case, the translation makes up the `nuisance' (confound) parameter a classifier must decode to enable accurate separation of the classes. Note that we have used the translation case as an example here, and the model specified above allows for more complex classes to be created. We note that since $h\in\mathbb{H}$ is monotonic and differentiable, its inverse $h^{-1}$ exists and is also differentiable.

We now describe the main Theorem of this paper clarifying the linear separation properties of the newly proposed CDT.

%

\begin{theorem}
\label{mainLemma}
\textbf{Linear Separability Theorem in CDT Space}  Let  $\mathbb{P}, \mathbb{Q}, \mathbb{H}$ follow be defined according to Definition \ref{def:model}. In addition, \textcolor{black}{let $h \in  \mathbb{H}$ satisfy the following conditions}:
\begin{enumerate}[i)]
\item  $\forall h \in \mathbb{H}, \;\;  h^{-1} \in \mathbb{H}$.

\item  $ \forall h \in \mathbb{H} \text{ and } \alpha_i>0 \text{ that satisfies } \sum_i \alpha_i = 1$, $h_\alpha^{-1} = \sum_i \alpha_i h_i^{-1} \in \mathbb{H}$.

\item $\forall h_{1}, h_{2} \in \mathbb{H}, \;\;  h_{1} \circ h_{2} \in \mathbb{H}$.

\end{enumerate}
Then the corresponding sets in the CDT space $\widehat{\mathbb{P}}, \widehat{\mathbb{Q}}$ are linearly separable.
\end{theorem}
We note that the linear separability theorem is independent of the choice of the reference $I_0$.  For a proof, see  \ref{appendix:mainLemma}.

\section{\textbf{Computational Algorithm}}
\label{CDTcomputation}
We now describe a numerical method for approximating the CDT given discrete data. Recall that the CDT is defined for continuous-time functions in contiguous, finite domain. In order to compute the CDT for a discrete-time signal, we need a way of estimating its cumulative function at any arbitrary coordinate. We do so via interpolation. Given a discrete signal of $N$ points and an interpolating model, the forward CDT can be estimated numerically  at all $N$ points. Our numerical method is designed when the reference function is $I_0(x) = 1$  for $x \in [0,1]$ (recall the linear separation properties of the CDT are independent of the choice of reference). The computation is formulated with the aid of B-splines \cite{Unser1993}. We use the B-spline of degree zero which guarantees that the reconstructed signals are always positive, which yields a low complexity algorithm ($O(N)$). \textcolor{black}{We note that under the specific construction below the approximated density functions will be discontinuous at the half way point between sampled nodes, and, as stated above, reconstruction at these points is not possible.}

 Let $\pi(x)$ be the  B-spline of degree zero of width $r$
\begin{align*}
\pi(x) &=
\begin{cases}
1  \text{ } &x \in [-\frac{1}{2}r, \frac{1}{2}r] \\
0 &\text{ elsewhere}
\end{cases}
\end{align*}
and define $ \Pi(x) = \int_{-\infty}^x  \pi(\tau)d\tau $ as
\begin{align}
\label{eq:Phi}
\Pi(x) &=
\begin{cases}
0 \text{ } &x < -\frac{1}{2}r\\
x+\frac{1}{2}r \text{ }&x \in [-\frac{1}{2}r, \frac{1}{2}r] \\
r \text{ } &x > \frac{1}{2}r.
\end{cases}
\end{align}
Let's denote a $N$-point discrete-time signal as ${\bf c} = [c_1, \cdots, c_N]$ and $x_i$ as the $i^{th}$ sample location of ${\bf c}$, i.e. ${\bf c}(x_i) = c_i$, $\forall i = 1,\cdots,N$.
We interpolate the discrete-time signal ${\bf c}$ with the B-spline of degree zero to be a continuous-time signal such as $I_1(x) = \sum_{i=1}^{N} c_i\pi(x-x_i)$ \textcolor{black}{for $x \in [x_1 - \frac{1}{2}r, x_N + \frac{1}{2}r]$}. Rewriting \eqref{eq:intcdt}, we have
\begin{align}
\label{eq:interp}
&\int_{x_1 -\frac{1}{2}r}^{f_1(x)} I_1(\tau) d\tau =\int_{x_1-\frac{1}{2}r}^{f_1(x)} \sum_{i=1}^{N} c_i\pi(\tau-x_i)d\tau =x
\end{align}
which can be simplified further by interchanging the sum and the integral, and then using $\Pi$ to denote the cumulative integral function of $\pi$ as
\begin{align}
\label{eq:interp2}
 \sum_{i=1}^{N} c_i\Pi(f_1(x) - x_i )= x.
\end{align}
By substituting \eqref{eq:Phi} into \eqref{eq:interp2} and taking the inverse of $\Pi(x)$ which is piecewise linear, $f_1(x)$ is computed according to the following algorithm:
\begin{enumerate}
\item When $0 < x < rc_1$, we have $c_1 \Pi( f_1(x)- x_1) = x$. Thus,
\[
f_1(x) = \frac{x}{c_1}+x_1-\frac{1}{2}r
\]

\item When $ -rc_n+  \sum_{i=1}^{n-1} c_i < x < rc_n+  \sum_{i=1}^{n-1} c_i$, we have $\sum_{i=1}^{n} c_i\Pi(f_1(x) - x_i ) = x$. Thus,
\begin{align}
\label{eq:comp}
f_1(x) = \frac{x- \sum_{i=1}^{n-1} c_i}{c_n} + x_n - \frac{1}{2}r.
\end{align}

\item Proceed until $n=N$. 

\end{enumerate}

\section{\textbf{Computational experiments}}
\label{CDTexamples}
In this section, we experimentally evaluate the properties of the CDT by comparing linear classification performed in CDT space with that in original signal space ($L^2$). Specifically, we investigate five cases of signal classification: classification of texture images from histograms, classification of accelerometer signals, classification of flow cytometry data,  classification of histograms from hand gesture image data, and classification of cell images from orientation histograms. Note that our goal is not to propose the ultimate, or optimal, classification method for each application, but rather to experimentally validate Theorem \ref{mainLemma} using both simulated (manufactured) data and diverse, real datasets. We note that, with the exception of the simulated case in Section \ref{sec:intro}, we have no precise knowledge of whether conditions $i)$, $ii)$, and $iii)$ for $\mathbb{H}$ specified in Theorem \ref{mainLemma} hold. Results seem to confirm, however, that the generative model specified in these conditions has a least some bearing on each problem investigated here.

As the CDT does not actually prescribe an optimal classifier, we quantify the degree of linear separability of the data by computing classification error using three different linear classifiers using a standard cross validation procedure (or leave-one-out cross validation when data size is small) that separates training and testing data. In addition, we provide more qualitative (visual) evidence, by computing a low dimensional projection of the data using training data only, that the CDT indeed tends to make data more linearly separable.

\subsection{Experimental procedure}

\begin{algorithm}[b!]
 \SetAlgorithmName{Algorithm}
1. Partition the dataset into 5 groups. Leave one fold out for testing and use the remaining fold for training.\\
\ForEach{training set}{
1. For SVM and PLDA, partition the training set into 5 groups again. Leave one fold out for validation and use remaining fold for training. For LDA, skip to step 2. \\
  \ForEach{training set (parameter sweep)}{
   1. Learn the classifier for different parameter values. \\
   2. Compute the validation error. \\
}
Return the best parameter of average validation error. \\
2. Learn the classifier with the optimal parameter. \\
3. Compute the testing error.\\
}
2. Compute the average classification error.
 \caption{5-fold cross validation}
 \label{algoSteps}
\end{algorithm}

Average classification error is compared using three different linear classifiers: Fisher's linear discriminant analysis (Fisher LDA) method \cite{Belhumeur1997}, the penalized LDA (pLDA) method of Wang et al \cite{Wang2011}, and the linear Support Vector Machine (SVM) method \cite{Vapnik1999}. All experiments were performed using the MATLAB \cite{MATLAB:2014b} programming language, while the SVM method was implemented using the LIBSVM package \cite{LIBSVM}. While the Fisher LDA method does not require parameter tuning, the linear SVM and pLDA methods require parameter tuning steps which were performed using $2^{nd}$ depth cross validation utilizing the training set only. In the SVM method, the parameter is set to reflect how much error the separating hyperplane is to tolerate, while the parameter in the pLDA method determine the regularization to be applied when computing the covariance matrix (refer to references \cite{Vapnik1999, Wang2011} for more details).

The low dimensional visualization plots were computed using the pLDA method, which in contrast to the standard LDA method can yield multi dimensional embeddings for the given data. The dimensions of each embedding are weighted according to a optimization metric, which combines a data separation term (given by LDA) and a `data fitting' term (given by the standard Principal Component Analysis cost function). For each experiment reported below, we utilize the pLDA method to visualize a 2-dimensional embedding of the \emph{testing} data. In each case, a subset of the data was used to estimate the lower dimensional embedding. Remaining (testing) data was used to obtain the visualizations. 

The computational experiments shown in Sections \ref{result:texture}, \ref{result:cyto}, \ref{result:hand}, \ref{result:HeLa} were computed using a five-fold cross validation strategy, with 80\% of the data used for training, and 20\% for testing. For experiment in Section \ref{result:accel}, due to small sample size, a leave-one-out cross validation is used instead. The experimental procedure is summarized in Algorithm \ref{algoSteps}. For more details on cross validation experimental procedures, refer to references \cite{hastie2009elements, bishop2006pattern}.

\begin{figure}[t!]
    \centering
    \subfloat[][PLDA projection in $L^2$ space]{
     \scalebox{0.35}{
%
%
\definecolor{mycolor1}{rgb}{0.00000,0.44700,0.74100}%
\definecolor{mycolor2}{rgb}{0.85000,0.32500,0.09800}%

\newcommand\mtlarge{\fontsize{20pt}{24pt}\selectfont}

\begin{tikzpicture}

\begin{axis}[%
width=6.027778in,
height=4.754167in,
at={(1.011111in,0.641667in)},
scale only axis,
every outer x axis line/.append style={black},
every x tick label/.append style={font=\color{black}},
xmin=-0.3,
xmax=0.2,
every outer y axis line/.append style={black},
every y tick label/.append style={font=\color{black}},
ymin=-0.6,
ymax=0.1,
axis x line*=bottom,
axis y line*=left,
label style={font=\mtlarge},   
xlabel={$1^{st}$ Discriminant Direction},
ylabel={$2^{nd}$ Discriminant Direction}, 
ticks=none
]
\addplot[only marks,mark=*,mark options={},mark size=1.5000pt,color=mycolor1] plot table[row sep=crcr,]{%
-0.190795818819149	-0.231871637970249\\
-0.176361706027135	-0.232244273365857\\
-0.174300583905253	-0.233711377299543\\
-0.17790690521398	-0.235386548143256\\
-0.174430525675975	-0.240571805880339\\
-0.172471002932898	-0.2397233498487\\
-0.173734001435501	-0.238725152638108\\
-0.175459746940213	-0.237696733828029\\
-0.106850917164976	-0.11837298260418\\
-0.0976283811615098	-0.116220029447537\\
-0.0887916135359397	-0.115951578108974\\
-0.0850914873794831	-0.116777977217959\\
-0.0784232596564768	-0.125286820323548\\
-0.0826893806741561	-0.123542123891696\\
-0.0782873402993915	-0.121094257176125\\
-0.0780547366789457	-0.119211488571202\\
0.00334807270162674	-0.0146023926605506\\
-0.0077935256118174	0.00536197964347423\\
-0.00438551830777217	0.00765560682954634\\
-0.000716356820121461	0.00838011885824825\\
0.0170329413563591	0.00110549501669848\\
0.0123206924803732	0.00354872623996714\\
0.0171924542624232	0.00652410446731916\\
0.0105037326966221	0.00768773198757917\\
0.023644814089809	-0.0500579147377905\\
0.0169343901605699	-0.00960556380122372\\
-0.0158136608985144	0.00696084977169542\\
-0.0103160111419456	0.00919767092634405\\
0.0247503445318247	0.00662375515823406\\
0.0196835765513778	0.00922744469029135\\
0.00869891680126089	0.010041515491793\\
-0.00468170724414261	0.00984711345869658\\
0.068824513420802	-0.249179197212935\\
0.0234381275684673	-0.0486554322059433\\
0.0145053760251581	-0.00906314403636065\\
0.00203312655362355	0.0107595413748176\\
0.0187848540278417	0.010652244954443\\
-0.00734995403054362	0.0109383432646965\\
-0.00773720528627698	0.0124552758915762\\
-0.00838249482356711	0.012523924976533\\
0.11934503135334	-0.376889042331661\\
0.0740831228168881	-0.239740656508486\\
0.0106357341876232	-0.0506204430031263\\
0.00284049953179097	-0.00922287158500669\\
-0.000945670707050673	0.0134179376407986\\
-0.0113691763867739	0.0139231845856378\\
-0.00494288562614974	0.0143485433490215\\
-0.0169739874167298	0.0121321345637873\\
0.158296313497745	-0.462361252712168\\
0.12219927365157	-0.376831827670649\\
0.0749184075434781	-0.251074519702468\\
0.0207409535456397	-0.0510132498226594\\
-0.00104810625680709	0.0158856336264888\\
-0.0109773585684199	0.0156018176456313\\
-0.0111282872682735	0.0133357702651908\\
-0.00795271463736242	0.0112435491307691\\
0.193883742803413	-0.582009107413739\\
0.162915102266962	-0.461721356388567\\
0.131521304871241	-0.377151545744774\\
0.0809083407683222	-0.241631682982642\\
0.0039985532589814	0.017852098020605\\
-0.0351667012148785	0.0121243762623034\\
0.0025762138275276	0.0124716960664397\\
0.00822561968865355	0.00916593094558988\\
};
\addplot[only marks,mark=*,mark options={},mark size=1.5000pt,color=mycolor2] plot table[row sep=crcr,]{%
-0.22688158340423	-0.289898362084748\\
-0.224494272658224	-0.29127587364625\\
-0.220492399530381	-0.291985688708699\\
-0.223389477198223	-0.292963662211701\\
-0.224148186341017	-0.296023195796046\\
-0.224150087952258	-0.29563579755484\\
-0.223875524964892	-0.294913440853956\\
-0.222782855598226	-0.29423582106037\\
-0.145998039729482	-0.173673881827495\\
-0.137609771019568	-0.174062091982736\\
-0.130363122911921	-0.174685828962033\\
-0.133960215972309	-0.176235487074064\\
-0.133271904496793	-0.182666010312478\\
-0.131991658160427	-0.181567060861425\\
-0.127563734172239	-0.17967711546166\\
-0.128271932041104	-0.178262544702045\\
-0.0900460575484744	-0.0928232302235582\\
-0.0821129567770062	-0.0916543895115091\\
-0.0790123262353304	-0.0920431829825079\\
-0.072550919080267	-0.0924537323782274\\
-0.0619161655234064	-0.100653538608315\\
-0.0628081959301227	-0.098673281478201\\
-0.0579659778236383	-0.0961179884497356\\
-0.0656619050421606	-0.0948552111442653\\
-0.0336634531924519	-0.0531308536680094\\
-0.0378246052402083	-0.0300532240442441\\
-0.0400518847866549	-0.0295301411336407\\
-0.034007499737011	-0.0287685733129615\\
-0.0102831716539949	-0.0357313863324632\\
-0.013021892497483	-0.0329624475736121\\
-0.013108075970289	-0.0303501427673226\\
-0.0255589077947535	-0.0296322028304519\\
0.023099147788825	-0.108362331160077\\
-0.00297759118770007	-0.0172076507589635\\
-0.0188406916992771	0.00210897924434841\\
-0.0181983595844977	0.00371422840364011\\
0.019188153288084	0.00129671198221893\\
0.0037320891920274	0.00357812042748837\\
-0.000601244394274133	0.00506799814264317\\
-0.0123124637078625	0.00506391687987812\\
0.081228977996709	-0.260245903853091\\
0.0374968900252045	-0.100293054676346\\
-0.00104662270624023	-0.013266369598889\\
-0.010907500823622	0.00918030584658462\\
0.011520174793951	0.0115225734678011\\
-0.00441379820262839	0.0126021613040157\\
-0.0116602235568854	0.0127155055045257\\
-0.0161780940785693	0.0117489273163102\\
0.133470800112869	-0.410972697195621\\
0.0836012764627646	-0.251422466628976\\
0.0339715354677807	-0.100787004931655\\
0.00190555800220522	-0.01074632574637\\
-0.00155503589335806	0.0137768076256237\\
-0.0179618824563561	0.013643510308712\\
-0.0147765261936481	0.012848083480888\\
-0.0190711995871458	0.01079246015441\\
0.17312814296273	-0.527363254188423\\
0.135339926215247	-0.410834810849496\\
0.0842927177914966	-0.260135406991747\\
0.0340412459250456	-0.1014489456487\\
-0.00858791529449106	0.0151153522480946\\
-0.0257314441691573	0.0133130169232084\\
-0.0132518661327381	0.0118366915788799\\
-0.00957225765851006	0.00936668775139712\\
};
\end{axis}
\end{tikzpicture}%
}  
 \label{fig:texture-plda-l2}
            }
    \subfloat[][PLDA projection in CDT space]{ 
     \scalebox{0.35}{
%
%
\definecolor{mycolor1}{rgb}{0.00000,0.44700,0.74100}%
\definecolor{mycolor2}{rgb}{0.85000,0.32500,0.09800}%
\newcommand\mtlarge{\fontsize{20pt}{24pt}\selectfont}

\begin{tikzpicture}

\begin{axis}[%
width=6.027778in,
height=4.754167in,
at={(1.011111in,0.641667in)},
scale only axis,
every outer x axis line/.append style={black},
every x tick label/.append style={font=\color{black}},
xmin=-0.25,
xmax=0.05,
every outer y axis line/.append style={black},
every y tick label/.append style={font=\color{black}},
ymin=-5,
ymax=3,
axis x line*=bottom,
axis y line*=left,
label style={font=\mtlarge},   
xlabel={$1^{st}$ Discriminant Direction},
ylabel={$2^{nd}$ Discriminant Direction}, 
ticks=none
]
\addplot[only marks,mark=*,mark options={},mark size=1.5000pt,color=mycolor1] plot table[row sep=crcr,]{%
-0.24559021858104	-3.09944310898847\\
-0.231592283314514	-3.44642503788163\\
-0.219398083965218	-3.70939631354277\\
-0.20881265777964	-3.91000470074272\\
-0.157408534598479	-4.65460382561586\\
-0.167282743844588	-4.5092926794961\\
-0.17825705111958	-4.36208570417071\\
-0.188663031368449	-4.21726709278544\\
-0.230298647131303	-1.80596696983581\\
-0.217008119487813	-2.33578516665461\\
-0.208538784547241	-2.73847869567676\\
-0.20249202857261	-3.04601336604963\\
-0.158158921166573	-4.18969145215998\\
-0.168190649970424	-3.96616735657706\\
-0.17801438646811	-3.74130597717844\\
-0.187444257999094	-3.51921043303072\\
-0.225762419520324	-0.530361345438979\\
-0.210343165007852	-1.23652476729795\\
-0.202134461573984	-1.7767679120691\\
-0.197699570910044	-2.18858317730057\\
-0.157566844560127	-3.73079296679791\\
-0.168758903719584	-3.42846004162056\\
-0.179616806182839	-3.12554567831951\\
-0.18752767864007	-2.82478969566586\\
-0.178560870884835	0.408547696924469\\
-0.217886215484422	-0.365667825120146\\
-0.205201874026178	-1.009300041471\\
-0.198509221915467	-1.50772392244454\\
-0.163893694184642	-3.36198986878211\\
-0.177215241917961	-2.99921757668931\\
-0.184931947653682	-2.63442590043804\\
-0.189817606434901	-2.27397065991892\\
-0.225203470233969	0.96905590252581\\
-0.175166507074302	0.474653140523606\\
-0.217081023591718	-0.213816985661977\\
-0.203003913626807	-0.797046427207027\\
-0.171735261695228	-2.97551551741909\\
-0.183266700907791	-2.55001803264628\\
-0.186725053120671	-2.12469296263982\\
-0.19385665206243	-1.69843928943413\\
-0.234320374961318	1.19418106919668\\
-0.220837530718528	0.942461886560657\\
-0.178704110928509	0.494341531322529\\
-0.210779043919792	-0.116922796810278\\
-0.17851450929327	-2.60672501942216\\
-0.183713221003806	-2.12041856347327\\
-0.18968745138516	-1.632078309061\\
-0.194571403978341	-1.14250722336653\\
-0.171723912128147	1.30777756378244\\
-0.2312776514345	1.14692476549989\\
-0.225337148831001	0.933440806395368\\
-0.183018495755001	0.544536665337269\\
-0.183404994618641	-2.22761641170218\\
-0.190349824736115	-1.6763582264628\\
-0.196195154086936	-1.12034902771485\\
-0.20279668356999	-0.56975165438558\\
-0.187990485709432	1.08834570120615\\
-0.182093098061477	1.2443062531809\\
-0.233806222300808	1.10606652570765\\
-0.223313066605165	0.918796306253684\\
-0.163095028630076	-1.89103245483183\\
-0.166333993255838	-1.28285872106387\\
-0.172483033552635	-0.665756389169399\\
-0.180035288642064	-0.0534290680426071\\
};
\addplot[only marks,mark=*,mark options={},mark size=1.5000pt,color=mycolor2] plot table[row sep=crcr,]{%
-0.0729483872082783	-3.80043097059453\\
-0.0753522886979437	-4.046824874026\\
-0.0785352231601146	-4.23453501674353\\
-0.0790622204456027	-4.37684605145481\\
-0.086120686731491	-4.90871078094699\\
-0.0840913308074386	-4.80635852476162\\
-0.0843915594848405	-4.70036929923099\\
-0.0823619354086848	-4.59611838773294\\
-0.0817299317142731	-2.60643143611586\\
-0.0906331173822523	-3.02551475962208\\
-0.09471928202134	-3.3429055193668\\
-0.094121348554927	-3.58258442194289\\
-0.0958112054132935	-4.48053512965037\\
-0.0948756019244652	-4.30674117418342\\
-0.0944976025054046	-4.12804765820665\\
-0.0954289512564763	-3.95423981010833\\
-0.054561343802085	-1.09805273634767\\
-0.0650797239099248	-1.72943792123884\\
-0.0755182688950105	-2.20894230507472\\
-0.0808169594807181	-2.57459115094684\\
-0.0889360921127389	-3.93686493445843\\
-0.088086091254775	-3.66966102048626\\
-0.0855457144804567	-3.40284247970244\\
-0.0872135855007804	-3.13582282731684\\
-0.0443275119766719	0.414753708694139\\
-0.0524178655143881	-0.429058443786093\\
-0.062998185205061	-1.0699774208852\\
-0.0708967052758941	-1.56122291962826\\
-0.086511256258643	-3.39030659703446\\
-0.0871009176409814	-3.03383871702633\\
-0.0847408331430207	-2.67454321211416\\
-0.083052848771496	-2.31806163935063\\
0.0148574807965427	1.42001141045726\\
-0.032951736682699	0.589116204370046\\
-0.0408415347306289	-0.174495028270194\\
-0.0525645809314526	-0.766776630133566\\
-0.0806148722270154	-2.96211400001752\\
-0.0781903821272703	-2.53400377914497\\
-0.0761357348123735	-2.1033040262475\\
-0.0683889920063014	-1.67497670250118\\
-0.0891683326222547	1.81733705433628\\
-0.0104236552267921	1.34681159952986\\
-0.0523450484093842	0.630113293968359\\
-0.0590215624988219	-0.0477306766057554\\
-0.0894092806975173	-2.57501936363518\\
-0.0886853297241774	-2.08135644430768\\
-0.0797853410404907	-1.5853508116351\\
-0.0733647403022642	-1.09286158339207\\
-0.0823738160840581	2.02621413366566\\
-0.0955593104826658	1.67015911217297\\
-0.0300796824153449	1.26766108160852\\
-0.0628715935210494	0.639177193871857\\
-0.0961664493687843	-2.20402205163012\\
-0.0897957898885584	-1.64956659655363\\
-0.082327610234984	-1.08967203853268\\
-0.0754005309855742	-0.533833891153327\\
-0.0782978311970875	1.93199766746333\\
-0.0934100526333766	1.85535956109055\\
-0.107993268716805	1.56480131398216\\
-0.0458344851324114	1.22396043359052\\
-0.101638372377816	-1.82159105843808\\
-0.0946868078847517	-1.20208660473813\\
-0.0884790495259133	-0.577215719743216\\
-0.0817062355429531	0.0420826597192572\\
};
\end{axis}
\end{tikzpicture}%
}  

             \label{fig:texture-plda-cdt}
             }
              \caption{PLDA projection for texture dataset}
  \label{fig:texture-plda}
\end{figure}
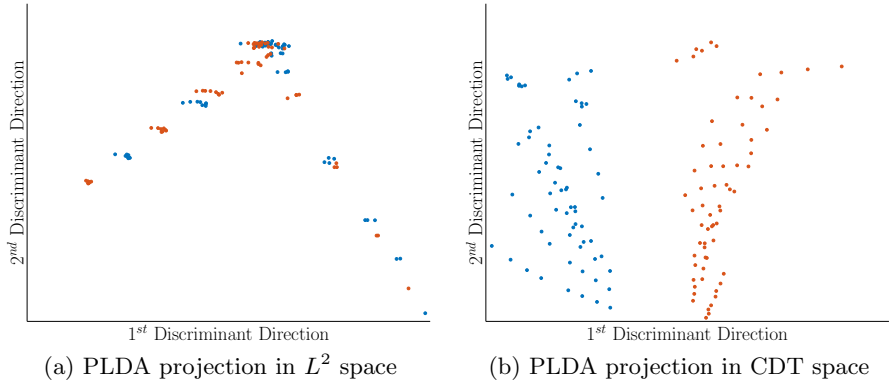

\subsection{Texture classification from intensity histograms}
\label{result:texture}
In this application, already discussed in the introduction as a motivating example, our goal is to utilize the CDT to distinguish between two types of texture images, under brightness and contrast variations, from their intensity histograms. Consider the textures displayed in the middle rows of \figref{fig:texture}. Their corresponding histograms are shown directly under and above each image, with variations in brightness and contrast. Variations in brightness correspond to translations in the histograms, while variations in contrast correspond to scalings (dilations) of the histograms. We note that such variations (translation and scaling) satisfy the necessary properties described in Theorem \ref{mainLemma}. Our theory thus predicts that the histogram data would be perfectly separable in CDT space. For testing this hypothesis, we generated a set of 128 images (2 sets of 64 images) by applying 8 random variations in brightness, with the translations in the range of $[0, 0.5]$, and 8 random variations in contrast, with scalings in the range of $[0.6, 1.67]$. Results are shown in \tableref{table:texture}, and confirm that 1) the data is not linearly separable in histogram space and 2) becomes linearly separable in CDT space. The lower dimensional representation of the original data using Penalized LDA also confirms this (see \figref{fig:texture-plda}).

\subsection{Activity Recognition with Accelerometer Data}
\label{result:accel}
\begin{figure}[t!]
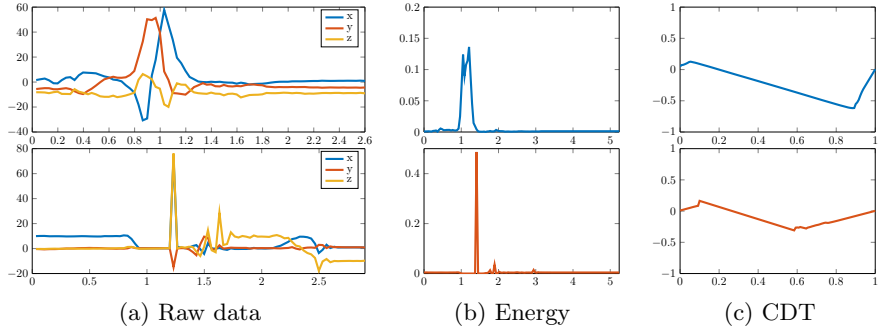

     \centering
     \subfloat[][Raw data]{ %
     \scalebox{0.4}{\input{input/accel-raw.tex}}  
\label{fig:accel-raw}}
     \subfloat[][Energy]{  %
     \scalebox{0.4}{\input{input/accel-energy.tex}}  
  \label{fig:accel-energy}}
     \subfloat[][CDT]{  %
     \scalebox{0.4}{\input{input/accel-cdt.tex}}  
  \label{fig:accel-cdt}}
     \caption{Two classes of accelerometer dataset, swinging (top row) vs free falling (bottom row)} 
     \label{fig:accel}
\end{figure}

An accelerometer is a device that records the acceleration of a moving object. Modern `smartphones' are commonly equipped with a 3-axis accelerometer that keeps track of the acceleration in 3 different directions $x$, $y$, and $z$, and accelerometers have been widely adapted to various wearable devices (e.g. watches) for human activity recognition. In this example, we aim to detect (classify) two different activities given accelerometer data obtained from an iphone 5. Class 1 consists of a person swinging arms while holding the phone. Class 2 consists of a phone being dropped to the ground. \figref{fig:accel-raw} shows the raw data recorded from the accelerometer for both cases. We note that in this case, the signals varied in length given the different duration of the episodes. Signals were zero-padded so that they match the length of the largest signal. 10 sample signals were acquired for each class. For each instance, the Energy =  $x^2 + y^2 + z^2$ is computed from the tri-axis measurements (see \figref{fig:accel-energy}). Here we compare the ability of the linear classification in original (energy) signal space versus in CDT space.

\begin{figure}[t!]
     \centering
     \subfloat[][PLDA projection in $L^2$ space]{ %
     \scalebox{0.3}{
%
%
\definecolor{mycolor1}{rgb}{0.00000,0.44700,0.74100}%
\definecolor{mycolor2}{rgb}{0.85000,0.32500,0.09800}%

\newcommand\mtlarge{\fontsize{20pt}{24pt}\selectfont}
\begin{tikzpicture}

\begin{axis}[%
width=6.027778in,
height=4.754167in,
at={(1.011111in,0.641667in)},
scale only axis,
every outer x axis line/.append style={black},
every x tick label/.append style={font=\color{black}},
xmin=-0.12,
xmax=0.02,
xticklabel style={
        /pgf/number format/fixed,
        /pgf/number format/precision=2
},
max space between ticks = 50,
try min ticks = 8,
every outer y axis line/.append style={black},
every y tick label/.append style={font=\color{black}},
ymin=-0.2,
ymax=0.25,
axis x line*=bottom,
axis y line*=left,
label style={font=\mtlarge},   
xlabel={$1^{st}$ Discriminant Direction},
ylabel={$2^{nd}$ Discriminant Direction}, 
ticks=none
]
\addplot[only marks,mark=halfdiamond*,mark options={},mark size=6.5000pt,color=mycolor1] plot table[row sep=crcr,]{%
-0.0974250617630781	0.153080289999584\\
-0.100444093058182	0.0400872926487463\\
0.0137329773807263	-0.0253237654870557\\
-0.0967099187807693	-0.0333292134430903\\
-0.0988397798161525	-0.0963775431682655\\
-0.100531301298005	0.0525202525786011\\
-0.0987372732327758	-0.198947219371946\\
-0.10129779683996	0.162194344682217\\
-0.0992759718853704	0.175452473655128\\
0.0136575087046689	-0.0041739706319799\\
};
\addplot[only marks,mark=10-pointed star,mark options={},mark size=6.5000pt,color=mycolor2] plot table[row sep=crcr,]{%
0.0192737108005679	0.0497615130034286\\
-0.0104449969814131	0.00756278304029311\\
0.0170324382397999	0.104633386148829\\
0.0180422113426467	0.0252922603720962\\
0.0150098487729725	0.00307898500832809\\
0.0187718774503026	-0.0889404806315097\\
0.0181992781796941	-0.0853203760656414\\
0.0181396093119285	0.214279572893312\\
0.0155395162576972	0.0403290379552695\\
-0.0815209702968916	-0.0320438779629028\\
};
\end{axis}
\end{tikzpicture}%
}  
\label{fig:accel-plda-l2}}
     \subfloat[][PLDA projection in CDT space]{  %
     \scalebox{0.3}{
%
%
\definecolor{mycolor1}{rgb}{0.00000,0.44700,0.74100}%
\definecolor{mycolor2}{rgb}{0.85000,0.32500,0.09800}%
\newcommand\mtlarge{\fontsize{20pt}{24pt}\selectfont}
\begin{tikzpicture}

\begin{axis}[%
width=6.027778in,
height=4.754167in,
at={(1.011111in,0.641667in)},
scale only axis,
every outer x axis line/.append style={black},
every x tick label/.append style={font=\color{black}},
xmin=-0.7,
xmax=0,
xticklabel style={
        /pgf/number format/fixed,
        /pgf/number format/precision=2
},
max space between ticks = 50,
try min ticks = 8,
every outer y axis line/.append style={black},
every y tick label/.append style={font=\color{black}},
ymin=-2,
ymax=1.5,
axis x line*=bottom,
axis y line*=left,
label style={font=\mtlarge},   
xlabel={$1^{st}$ Discriminant Direction},
ylabel={$2^{nd}$ Discriminant Direction}, 
ticks=none
]
\addplot[only marks,mark=halfdiamond*,mark options={},mark size=6.5000pt,color=mycolor1] plot table[row sep=crcr,]{%
-0.680405131407784	0.945639380122899\\
-0.680323381326483	0.627817320782706\\
-0.666513012928055	-0.621378742125285\\
-0.674733721652905	0.525294202913264\\
-0.677148413202729	-1.6748335093233\\
-0.679383414435781	1.01088757111567\\
-0.681278094555621	-0.416118741657684\\
-0.676789605584714	0.704248954566165\\
-0.681083562605907	0.691117161763947\\
-0.465102379774045	-1.82040803024515\\
};
\addplot[only marks,mark=10-pointed star,mark options={},mark size=6.5000pt,color=mycolor2] plot table[row sep=crcr,]{%
-0.089712537537503	-1.17128391501183\\
-0.338882513386851	0.483050149074447\\
-0.0890493206037153	1.00981204459301\\
-0.0892644294514309	0.110634093327241\\
-0.0783134419510565	-0.010655888627202\\
-0.0884447599551922	0.0204209544600755\\
-0.0840840621477934	0.434896704111561\\
-0.0862001532158563	1.08217720913554\\
-0.085938798992659	0.938869170916821\\
-0.0909499504562918	0.802548111190787\\
};
\end{axis}
\end{tikzpicture}%
}  
  \label{fig:accel-plda-l2}}
     \caption{PLDA projection for accelerometer dataset}
     \label{fig:accel-plda}
\end{figure}
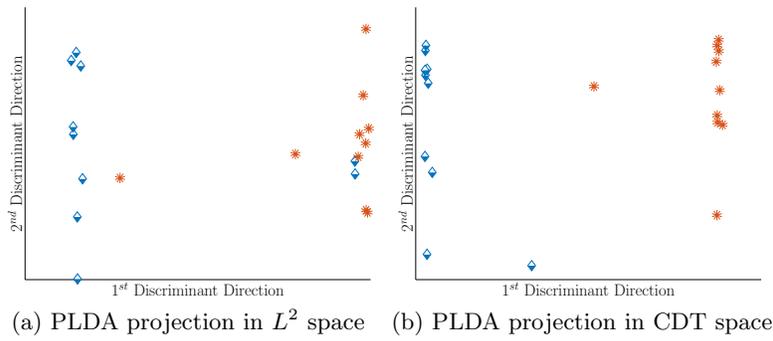

Results are shown in \tableref{table:accel}, and clearly indicates that the data becomes linearly separable in CDT space. The lower dimensional representation of the original data using Penalized LDA (PLDA) \cite{Wang2011} indicates (see \figref{fig:accel-plda}) that each class forms convex hulls that are linearly separable in CDT space but not in energy signal space. For this example, both training and testing data are represented in the lower dimensional embedding in \figref{fig:accel-plda}. By seeing  \figref{fig:accel-plda}, we can verify that the linear classifier computed using only training set correctly separates both training and testing set in CDT space, but not in energy signal space.

\begin{table}[t!]
\caption{Average classification error of the accelerometer dataset}
\centering
\begin{tabular}{|c|c|c|c|}
    \hline
   Classifier type&{Dataset} & $L^2$ space &  CDT space\\ 
    \hline
        \multirow{2}{*}{Fisher LDA}    &Training set & 0 \% & 0\% \\
    \hhline{~~~}  
    & Testing set & {\bf 50} \% &  {\bf 10}\% \\
      \hhline{----}  

    \multirow{2}{*}{PLDA}  &   Training set &  0\% &0 \% \\
    \hhline{~~~}  
    &Testing set & {\bf 60 } \% &{\bf 5} \% \\
      \hhline{----}  

    \multirow{2}{*}{Linear SVM} &  Training set &  8.75\% &7.5 \% \\
    \hhline{~~~}  
     &Testing set & {\bf 55 } \% &{\bf 10} \% \\
    \hline  
 \end{tabular}
 \label{table:accel}
\end{table}

In this experiment, it is apparent that the signals varied in terms of intensity and the location where the maximum peak has occurred, and this explains the inability of linear classifiers to perform well in original (energy) signal space. As explained above, the CDT is able to overcome such variations.

\begin{figure}[b!]
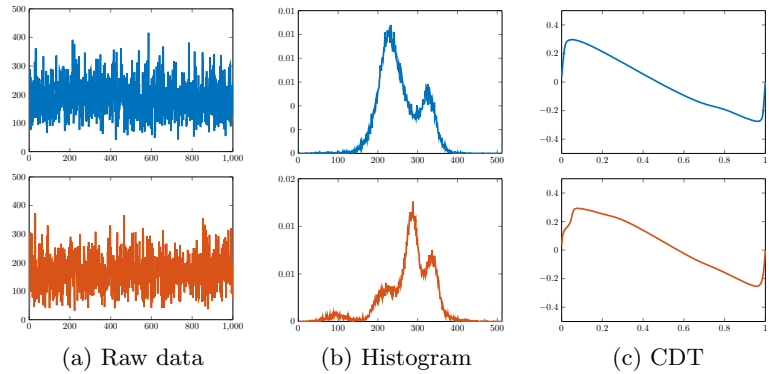

    \centering
    \subfloat[][Raw data]{
     \scalebox{0.32}{\input{input/cyto-raw.tex}}  
 \label{fig:cyto-raw}
            }
    \subfloat[][Histogram]{ 
     \scalebox{0.32}{\input{input/cyto-hist.tex}}  

             \label{fig:cyto-hist}
             }
     \subfloat[][CDT]{ 
     \scalebox{0.32}{\input{input/cyto-cdt.tex}}  

             \label{fig:cyto-cdt}
             }
  \caption{Two classes of flow cytometry data, AML (top row) vs. Normal (bottom row)}
  \label{fig:cyto}
\end{figure}

\begin{figure}[b!]
    \centering
    \subfloat[][PLDA projection in $L^2$ space]{
     \scalebox{0.35}{
%
%
\definecolor{mycolor1}{rgb}{0.00000,0.44700,0.74100}%
\definecolor{mycolor2}{rgb}{0.85000,0.32500,0.09800}%

\pgfplotsset{scaled x ticks=false}

\begin{tikzpicture}

\begin{axis}[%
width=6.027778in,
height=4.754167in,
at={(1.011111in,0.641667in)},
scale only axis,
every outer x axis line/.append style={black},
every x tick label/.append style={font=\color{black}},
xmin=-0.06,
xmax=0.06,
xticklabel style={
        /pgf/number format/fixed,
        /pgf/number format/precision=2
},
max space between ticks = 50,
try min ticks = 8,
every outer y axis line/.append style={black},
every y tick label/.append style={font=\color{black}},
ymin=-0.04,
ymax=0.08,
axis x line*=bottom,
axis y line*=left
]
\addplot[only marks,mark=halfdiamond*,mark options={},mark size=6.5000pt,color=mycolor1] plot table[row sep=crcr,]{%
-0.0287969623289152	0.00829350831004176\\
0.0174090866504028	0.0441052070573635\\
0.00266161942638086	0.0297418610651941\\
0.0111340528787224	0.0162742969619787\\
0.0338126958533513	0.00917279515105841\\
-0.0340077938513646	0.0064557125633773\\
0.0133784707945234	0.0156938596032529\\
-0.059614443249015	0.0288364277730185\\
};
\addplot[only marks,mark=10-pointed star,mark options={},mark size=6.5000pt,color=mycolor2] plot table[row sep=crcr,]{%
-0.00634763571355253	0.0382986830554683\\
-0.0153747037444061	0.0290842485730656\\
0.0201298805838201	0.0507037282166762\\
0.00722317498280659	0.037229820564118\\
-0.0201667983022694	0.0229284482037519\\
-0.000718818974063611	0.0347418350839634\\
0.00562582140403521	-0.0210095278170255\\
0.0209508363755758	0.0621120458895121\\
0.00229824619122652	0.038268917423074\\
-0.0417221254296401	0.00648705290730075\\
0.0355358982811853	0.0491059356055775\\
-0.000372171975661806	0.0259489496468361\\
-0.0332711303544618	0.0258005361135009\\
-0.0378535100242986	0.0154458674093153\\
-0.0158608799649674	0.0258436069993293\\
-0.0122347499637622	0.035363611309332\\
0.0129825460731747	0.0240139873418449\\
0.0024977768106472	0.0190794159338957\\
-0.0315318503908867	0.0163318714499436\\
0.011180313618846	0.0287012035801639\\
0.00971952340577557	0.0205240292779775\\
-0.0150610614403645	0.0211961641506833\\
-0.00527729979130038	0.0271292358009222\\
0.0016204268290707	0.0441315378445603\\
-0.0228964035393529	0.0494477026654453\\
0.0386797723942796	0.0223301571203684\\
0.0305048110672709	0.0228370894367813\\
-0.0288840525456507	0.0272853419661722\\
-0.01225659532596	0.0137270266447537\\
-0.0220632685910374	0.010039108298406\\
-0.0314030079759367	0.0700095614971402\\
-0.0314713143044981	0.0184117284917796\\
0.0110800146559929	0.0365132736348884\\
-0.0176978427080037	0.0473432726197518\\
-0.0167154495319363	0.0108612741854694\\
-0.0462367735500986	0.00827847039654808\\
-0.0169303413345482	0.00886206282903681\\
-0.0296144968268042	0.0127342405315545\\
-0.0229784167288787	0.0216253982050737\\
-0.0276238490927088	0.00383284660080567\\
-0.0352352617497619	0.0367922243600528\\
-0.00771973553584723	0.0396026757376848\\
0.00546917316937806	0.0429654761864224\\
-0.0269566266850842	0.0287784817272551\\
-0.0399861234972888	-0.0096636932832438\\
-0.00045845172823122	0.034456845846049\\
-0.015648524797167	0.02675168232629\\
0.0353540447158462	0.0129193217484573\\
-0.0109120448655419	0.062493066601921\\
0.021117919567903	0.018652443086418\\
-0.000819545964146434	0.0322855791499298\\
0.00396398797072464	0.0607667359002538\\
-0.00132444818265244	0.0193619136836071\\
0.0417620766588098	0.0286865865022676\\
0.0227055912866055	0.0305672174461554\\
};
\end{axis}
\end{tikzpicture}%
}  
 \label{fig:cyto-l2plda}
            }
    \subfloat[][PLDA projection in CDT space]{ 
     \scalebox{0.35}{
%
%
\definecolor{mycolor1}{rgb}{0.00000,0.44700,0.74100}%
\definecolor{mycolor2}{rgb}{0.85000,0.32500,0.09800}%
\begin{tikzpicture}

\begin{axis}[%
width=6.027778in,
height=4.754167in,
at={(1.011111in,0.641667in)},
scale only axis,
every outer x axis line/.append style={black},
every x tick label/.append style={font=\color{black}},
xmin=0,
xmax=1.2,
xticklabel style={
        /pgf/number format/fixed,
        /pgf/number format/precision=2
},
max space between ticks = 50,
try min ticks = 8,
every outer y axis line/.append style={black},
every y tick label/.append style={font=\color{black}},
ymin=-4.5,
ymax=-0.5,
axis x line*=bottom,
axis y line*=left
]
\addplot[only marks,mark=halfdiamond*,mark options={},mark size=6.5000pt,color=mycolor1] plot table[row sep=crcr,]{%
0.619493196533741	-1.47301920057812\\
0.782787167482816	-3.20250216575276\\
0.661942470424939	-2.95824323608387\\
0.934088341705501	-1.35646325768684\\
0.849831914154859	-3.12772069915729\\
0.65732908215025	-1.55585025068487\\
0.790216553033491	-0.621453085463216\\
0.383786706333788	-3.19241149734756\\
};
\addplot[only marks,mark=10-pointed star,mark options={},mark size=6.5000pt,color=mycolor2] plot table[row sep=crcr,]{%
0.851127741166455	-1.16003355466263\\
0.712735127446467	-2.21751883706343\\
0.280077621043013	-4.02708470933492\\
0.913259643316257	-1.89184720181925\\
0.763277752704832	-1.40640692363104\\
0.864568321168412	-2.60833481321276\\
0.0316950173385695	-1.67760529282101\\
0.251512646117082	-4.11553650195605\\
0.526353723075498	-3.21085851774001\\
0.674855081119742	-2.30743734224067\\
0.563814958440382	-3.90069759876779\\
0.681258833402722	-2.37328048489655\\
0.669137774371859	-2.36683821375567\\
0.531975357601297	-2.20144224928784\\
0.800335202742886	-1.3462013861052\\
0.606800198752634	-2.39559679087642\\
0.891829961922167	-2.01037583818678\\
0.795958507823975	-2.21142532858591\\
0.712032322191888	-1.28923269535444\\
0.702174992857948	-2.93977433029553\\
0.698906558878889	-2.40872207333248\\
0.637635257948422	-1.75337096024413\\
0.833413760829532	-1.83081608838727\\
0.767811377299766	-2.71575548566831\\
0.746885195584998	-2.79243097986594\\
0.69661800230727	-3.68395278306899\\
0.879650289341061	-3.13367895018343\\
0.680551854980869	-2.15884501673532\\
0.633825120689727	-1.99837941706769\\
0.639527668423438	-1.48921429811448\\
0.801768119721106	-3.08986424394824\\
0.644368201759704	-1.86678126770946\\
0.731963800635333	-2.78277056443679\\
0.384069538001654	-3.40842668609812\\
0.570855892145645	-0.859854183435423\\
0.879986611268049	-0.944006453399733\\
0.626381188429383	-1.26965418727041\\
0.621530285320493	-1.24636307801836\\
0.691995149542098	-0.98207558257113\\
0.420855918444547	-2.27398946163259\\
0.757376674298468	-2.40624449553674\\
0.831078351269191	-2.45175241757805\\
0.939763674857005	-2.75857160494299\\
0.629861881444098	-2.57898277285985\\
0.451359469558748	-1.45468475061702\\
0.757371643788188	-2.56572897457738\\
0.765753121058277	-1.52964679536673\\
1.06915464968801	-0.867487313856498\\
0.463269433678365	-3.56588319268657\\
0.89160729432819	-2.81751782184805\\
0.487866908589528	-3.04346119289301\\
0.940774973717411	-2.4335651843765\\
0.608358891598101	-1.55566555588107\\
1.07262169540298	-2.00230257928945\\
0.821108478920522	-3.19239451550114\\
};
\end{axis}
\end{tikzpicture}%
}  

             \label{fig:cyto-cdtplda}
             }
    \caption{PLDA projection for flow cytometry dataset}
     \label{fig:cyto-plda}
\end{figure}
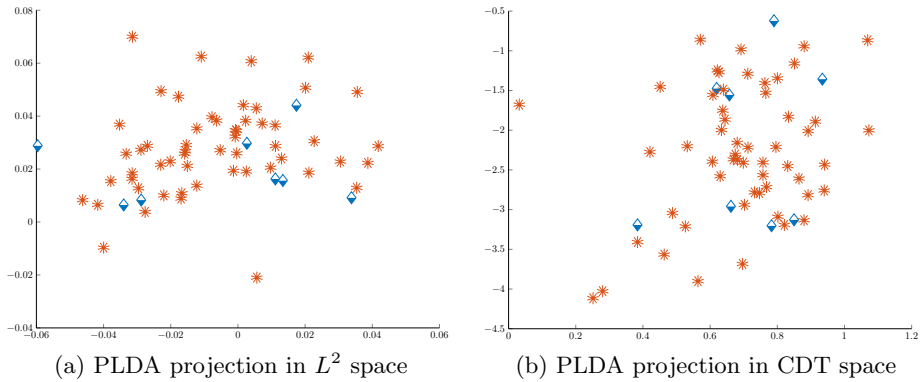

\begin{table}[b!]
\caption{Average Classification Error of the flow cytometry dataset}
\label{table:cyto}
\centering
\begin{tabular}{|c|c|c|c|}
    \hline
   Classifier type& Dataset & $L^2$ space  &  CDT space\\ 
    \hline
    \multirow{2}{*}{Fisher LDA}       
    &  Training set & 6.81 \% & 5.82\% \\
    \hhline{~~~}  
    &Testing set & {\bf 15.01 } \% &  {\bf 11.31} \% \\
    \hline
    \multirow{2}{*}{PLDA}       
    &  Training set & 11.55 \% & 7.75\% \\
    \hhline{~~~}  
    &Testing set & {\bf 12.03 } \% &  {\bf 9.15} \% \\
    \hline
    \multirow{2}{*}{Linear SVM}     
  &Training set & 10.39 \% & 8.65\% \\
    \hhline{~~~}  
     & Testing set & {\bf 11.46} \% &  {\bf 8.88}\% \\
    \hline  
 \end{tabular}
\end{table}

\subsection{Flow Cytometry}
\label{result:cyto}

Flow Cytometry is a technique used to analyze light emission properties of grouped cells using fluorescence markers. In this example, we utilize an existing database (the FlowRepository database \cite{dataset-flowcytometry}) to distinguish DNA histograms between normal subjects and donors diagnosed with acute myeloid leukemia (AML), obtained from peripheral blood or bone marrow aspirates. The data included 8 measurements per each subject, where fluorochrome signals were detected at the 620nm wavelength specifically. Sample data is shown in \figref{fig:cyto-raw}, where the x-axis represents each cell that passed through the flow cytometry sensor, and the y-axis correspond to the DNA intensity measurement of the cell at wavelength 620nm. The intensity histogram with 1024 intensity levels are computed and their corresponding CDTs (see \figref{fig:cyto-hist} and \figref{fig:cyto-cdt}).

The average classification error is reported in \tableref{table:cyto}. 
We note that the classification in the signal space using LDA (test accuracy of 84.99\%) is worse than the line of chance (87.5\%), given the uneven distribution of patient data. Comparison with the line of chance and the classification accuracy in histogram space using PLDA or SVM also suggests that linear classifiers trained are more or less equivalant to random classification. However, classifying data in CDT space suggests that linear separation is possible, and the Cohen's Kappa for this computation (0.3) confirms fair agreement \cite{altman1990practical}.

\subsection{Cambridge Hand Dataset}
\label{result:hand}
\begin{figure}[t!]
    \centering
    \subfloat[][Raw data]{
     \scalebox{0.3}{
%
%
\begin{tikzpicture}

\begin{axis}[%
width=2.545833in,
height=1.909375in,
at={(2.7625in,0.21772in)},
scale only axis,
axis on top,
separate axis lines,
every outer x axis line/.append style={black},
every x tick label/.append style={font=\color{black}},
xmin=0.5,
xmax=320.5,
tick align=outside,
every outer y axis line/.append style={black},
every y tick label/.append style={font=\color{black}},
y dir=reverse,
ymin=0.5,
ymax=240.5,
hide axis
]
\addplot [forget plot] graphics [xmin=0.5,xmax=320.5,ymin=0.5,ymax=240.5] {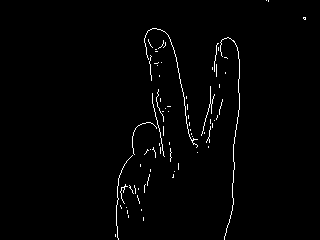};
\end{axis}

\begin{axis}[%
width=2.545833in,
height=1.909375in,
at={(0.108333in,0.21772in)},
scale only axis,
axis on top,
separate axis lines,
every outer x axis line/.append style={black},
every x tick label/.append style={font=\color{black}},
xmin=0.5,
xmax=320.5,
tick align=outside,
every outer y axis line/.append style={black},
every y tick label/.append style={font=\color{black}},
y dir=reverse,
ymin=0.5,
ymax=240.5,
hide axis
]
\addplot [forget plot] graphics [xmin=0.5,xmax=320.5,ymin=0.5,ymax=240.5] {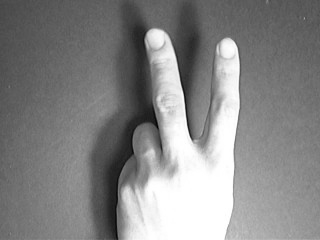};
\end{axis}

\begin{axis}[%
width=2.545833in,
height=1.909375in,
at={(2.7625in,4.636794in)},
scale only axis,
axis on top,
separate axis lines,
every outer x axis line/.append style={black},
every x tick label/.append style={font=\color{black}},
xmin=0.5,
xmax=320.5,
tick align=outside,
every outer y axis line/.append style={black},
every y tick label/.append style={font=\color{black}},
y dir=reverse,
ymin=0.5,
ymax=240.5,
hide axis
]
\addplot [forget plot] graphics [xmin=0.5,xmax=320.5,ymin=0.5,ymax=240.5] {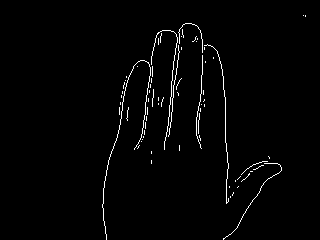};
\end{axis}

\begin{axis}[%
width=2.545833in,
height=1.909375in,
at={(2.7625in,2.427257in)},
scale only axis,
axis on top,
separate axis lines,
every outer x axis line/.append style={black},
every x tick label/.append style={font=\color{black}},
xmin=0.5,
xmax=320.5,
tick align=outside,
every outer y axis line/.append style={black},
every y tick label/.append style={font=\color{black}},
y dir=reverse,
ymin=0.5,
ymax=240.5,
hide axis
]
\addplot [forget plot] graphics [xmin=0.5,xmax=320.5,ymin=0.5,ymax=240.5] {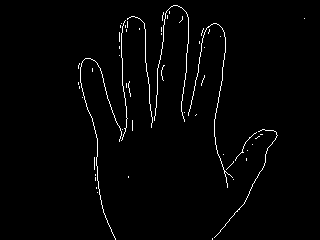};
\end{axis}

\begin{axis}[%
width=2.545833in,
height=1.909375in,
at={(0.108333in,2.427257in)},
scale only axis,
axis on top,
separate axis lines,
every outer x axis line/.append style={black},
every x tick label/.append style={font=\color{black}},
xmin=0.5,
xmax=320.5,
tick align=outside,
every outer y axis line/.append style={black},
every y tick label/.append style={font=\color{black}},
y dir=reverse,
ymin=0.5,
ymax=240.5,
hide axis
]
\addplot [forget plot] graphics [xmin=0.5,xmax=320.5,ymin=0.5,ymax=240.5] {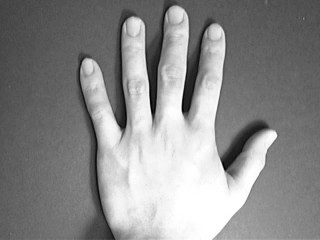};
\end{axis}

\begin{axis}[%
width=2.545833in,
height=1.909375in,
at={(0.108333in,4.636794in)},
scale only axis,
axis on top,
separate axis lines,
every outer x axis line/.append style={black},
every x tick label/.append style={font=\color{black}},
xmin=0.5,
xmax=320.5,
tick align=outside,
every outer y axis line/.append style={black},
every y tick label/.append style={font=\color{black}},
y dir=reverse,
ymin=0.5,
ymax=240.5,
hide axis
]
\addplot [forget plot] graphics [xmin=0.5,xmax=320.5,ymin=0.5,ymax=240.5] {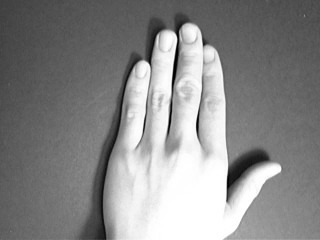};
\end{axis}
\end{tikzpicture}%
}  
 \label{fig:hand-im}
            }
    \subfloat[][Histogram]{ 
     \scalebox{0.3}{\input{input/hand-energy.tex}}  

             \label{fig:hand-energy}
             }
     \subfloat[][CDT]{
     \scalebox{0.3}{\input{input/hand-cdt.tex}}  

            \label{fig:hand-cdt}
             }             
            \caption{Three different classes of hand gestures dataset}
  \label{fig:hand}
\end{figure}

The Cambridge hand gesture dataset consists of 900 image sequences of 3 primitive hand shapes (see \figref{fig:hand-im}) where each image sequence consists of around 60 frames of 3 different motions \cite{dataset-hand}. In this example, we try to distinguish 3 different hand shapes; flat, spread, and v-shape. There are 2678 images for flat hands, 2992 images for spread hands, and 2764 images for v-shape hands, and each image was taken under arbitrary positioning and illumination. A preprocessing step computes the edge of each image (240 x 320 pixels large) and the corresponding indices of the edge pixels. Two histograms are created counting $x$ coordinates and $y$ coordinates of the edge pixels per image \figref{fig:hand-energy}. Corresponding CDTs are computed for each $x$ and $y$ histogram \figref{fig:hand-cdt}. The classification is done with concatenation of two $x$ and $y$ histograms and concatenation of two $x$ and $y$ CDTs. 

\begin{figure}[t!]
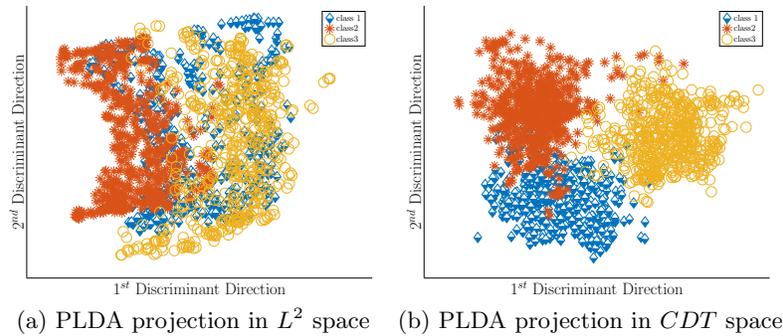

    \centering
    \subfloat[][PLDA projection in $L^2$ space]{
     \scalebox{0.3}{\input{input/hand-l2-plda-ts.tex}}  
 \label{fig:hand-l2plda}
            }
    \subfloat[][PLDA projection in $CDT$ space]{ 
     \scalebox{0.3}{\input{input/hand-cdt-plda-ts.tex}}  

             \label{fig:hand-cdtplda}
             }
    \caption{PLDA projection for hand gesture dataset}
     \label{fig:hand-plda}
\end{figure}

\begin{table}[h!]
\caption{Average Classification Error of the hand gestures dataset}
\centering
\begin{tabular}{|c|c|c|c|}
    \hline
   Classifier type& Dataset & $L^2$ space   &  CDT space\\ 
    \hline
    \multirow{2}{*}{Fisher LDA}       
  & Training set & 13.92 \% & 4.58\% \\
    \hhline{~~~}  
    & Testing set & {\bf 16.11} \% &  {\bf 5.76}\% \\
      \hline
          \multirow{2}{*}{PLDA}       
  & Training set & 38.02 \% &  6.73\% \\
    \hhline{~~~}  
    & Testing set & {\bf 38.21} \% &  {\bf 6.97}\% \\
      \hline
    \multirow{2}{*}{Linear SVM} 
  & Training set &  13.77\% & 1.27\% \\
    \hhline{~~~}  
     & Testing set & {\bf 15.73 } \% &  {\bf 1.65}\% \\
     \hline 
 \end{tabular}
 \label{table:hand}
\end{table}

Results are shown in \tableref{table:hand}, which clearly indicate that the data becomes more linearly separable in CDT space. As in previous examples, the two dimensional representation of the original testing data using Penalized LDA (PLDA) \cite{Wang2011} indicates (see \figref{fig:hand-plda}) that classes form convex hulls that are linearly separable in CDT space and not in histogram space. Moreover, this example shows that the CDT can be applied to multi-class problems which would enhance the simplicity of the classification problem.

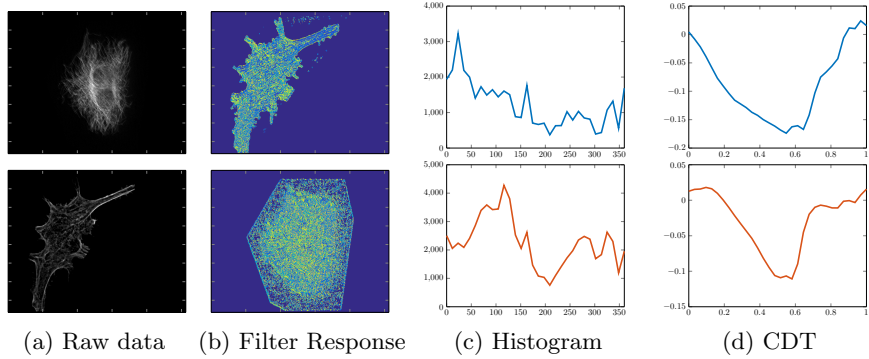
\begin{figure}[h!]
     \centering
     \subfloat[][Raw data]{ %
     \scalebox{0.3}{
%
%
\begin{tikzpicture}

\begin{axis}[%
width=3.1in,
height=2.479167in,
at={(0.172222in,3.0625in)},
scale only axis,
axis on top,
separate axis lines,
every outer x axis line/.append style={black},
every x tick label/.append style={font=\color{black}},
xmin=0.5,
xmax=512.5,
xtick={100,200,300,400,500},
xticklabels={\empty},
every outer y axis line/.append style={black},
every y tick label/.append style={font=\color{black}},
y dir=reverse,
ymin=0.5,
ymax=382.5,
ytick={50,100,150,200,250,300,350},
yticklabels={\empty}
]
\addplot [forget plot] graphics [xmin=0.5,xmax=512.5,ymin=0.5,ymax=382.5] {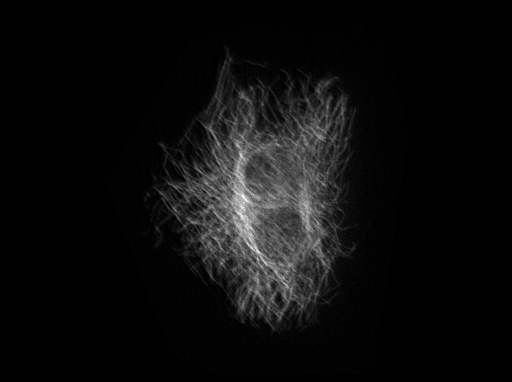};
\end{axis}

\begin{axis}[%
width=3.1in,
height=2.479167in,
at={(0.172222in,0.291667in)},
scale only axis,
axis on top,
separate axis lines,
every outer x axis line/.append style={black},
every x tick label/.append style={font=\color{black}},
xmin=0.5,
xmax=512.5,
xtick={100,200,300,400,500},
xticklabels={\empty},
every outer y axis line/.append style={black},
every y tick label/.append style={font=\color{black}},
y dir=reverse,
ymin=0.5,
ymax=382.5,
ytick={50,100,150,200,250,300,350},
yticklabels={\empty}
]
\addplot [forget plot] graphics [xmin=0.5,xmax=512.5,ymin=0.5,ymax=382.5] {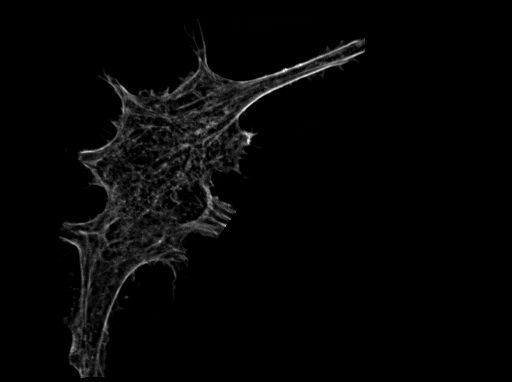};
\end{axis}
\end{tikzpicture}%
}  
\label{fig:hela-raw}}
     \subfloat[][Filter Response]{   %
     \scalebox{0.3}{
%
%
\begin{tikzpicture}

\begin{axis}[%
width=3.1in,
height=2.479167in,
at={(0.172222in,3.0625in)},
scale only axis,
axis on top,
separate axis lines,
every outer x axis line/.append style={black},
every x tick label/.append style={font=\color{black}},
xmin=0.5,
xmax=512.5,
xtick={100,200,300,400,500},
xticklabels={\empty},
every outer y axis line/.append style={black},
every y tick label/.append style={font=\color{black}},
y dir=reverse,
ymin=0.5,
ymax=382.5,
ytick={50,100,150,200,250,300,350},
yticklabels={\empty}
]
\addplot [forget plot] graphics [xmin=0.5,xmax=512.5,ymin=0.5,ymax=382.5] {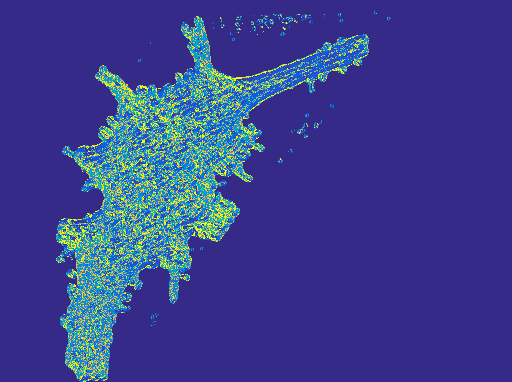};
\end{axis}

\begin{axis}[%
width=3.1in,
height=2.479167in,
at={(0.172222in,0.291667in)},
scale only axis,
axis on top,
separate axis lines,
every outer x axis line/.append style={black},
every x tick label/.append style={font=\color{black}},
xmin=0.5,
xmax=512.5,
xtick={100,200,300,400,500},
xticklabels={\empty},
every outer y axis line/.append style={black},
every y tick label/.append style={font=\color{black}},
y dir=reverse,
ymin=0.5,
ymax=382.5,
ytick={50,100,150,200,250,300,350},
yticklabels={\empty}
]
\addplot [forget plot] graphics [xmin=0.5,xmax=512.5,ymin=0.5,ymax=382.5] {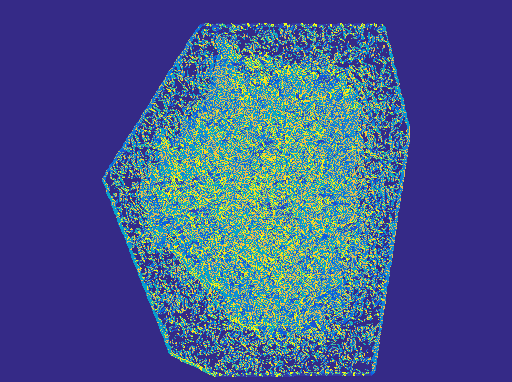};
\end{axis}
\end{tikzpicture}%
}  
  \label{fig:hela-gabor}}
     \subfloat[][Histogram]{   %
     \scalebox{0.3}{
%
%
\definecolor{mycolor1}{rgb}{0.00000,0.44700,0.74100}%
\definecolor{mycolor2}{rgb}{0.85000,0.32500,0.09800}
\begin{tikzpicture}

\begin{axis}[%
width=3.1in,
height=2.479167in,
at={(0.172222in,0.291667in)},
scale only axis,
separate axis lines,
every outer x axis line/.append style={black},
every x tick label/.append style={font=\color{black}},
xmin=0,
xmax=360,
every outer y axis line/.append style={black},
every y tick label/.append style={font=\color{black}},
ymin=0,
ymax=5000
]
\addplot [color=mycolor2,solid,line width=2.0pt,forget plot]
  table[row sep=crcr]{%
0	2503\\
11.6129032258065	2056\\
23.2258064516129	2239\\
34.8387096774194	2091\\
46.4516129032258	2408\\
58.0645161290323	2861\\
69.6774193548387	3392\\
81.2903225806452	3585\\
92.9032258064516	3421\\
104.516129032258	3442\\
116.129032258065	4273\\
127.741935483871	3802\\
139.354838709677	2527\\
150.967741935484	2056\\
162.58064516129	2619\\
174.193548387097	1466\\
185.806451612903	1080\\
197.41935483871	1035\\
209.032258064516	759\\
220.645161290323	1101\\
232.258064516129	1415\\
243.870967741935	1719\\
255.483870967742	1973\\
267.096774193548	2350\\
278.709677419355	2478\\
290.322580645161	2380\\
301.935483870968	1693\\
313.548387096774	1835\\
325.161290322581	2625\\
336.774193548387	2301\\
348.387096774194	1196\\
360	1953\\
};
\end{axis}

\begin{axis}[%
width=3.1in,
height=2.479167in,
at={(0.172222in,3.0625in)},
scale only axis,
separate axis lines,
every outer x axis line/.append style={black},
every x tick label/.append style={font=\color{black}},
xmin=0,
xmax=360,
every outer y axis line/.append style={black},
every y tick label/.append style={font=\color{black}},
ymin=0,
ymax=4000
]
\addplot [color=mycolor1,solid,line width=2.0pt,forget plot]
  table[row sep=crcr]{%
0	1936\\
11.6129032258065	2199\\
23.2258064516129	3223\\
34.8387096774194	2189\\
46.4516129032258	1998\\
58.0645161290323	1408\\
69.6774193548387	1727\\
81.2903225806452	1493\\
92.9032258064516	1644\\
104.516129032258	1442\\
116.129032258065	1606\\
127.741935483871	1504\\
139.354838709677	884\\
150.967741935484	860\\
162.58064516129	1762\\
174.193548387097	703\\
185.806451612903	665\\
197.41935483871	701\\
209.032258064516	374\\
220.645161290323	629\\
232.258064516129	633\\
243.870967741935	1024\\
255.483870967742	789\\
267.096774193548	1033\\
278.709677419355	848\\
290.322580645161	814\\
301.935483870968	395\\
313.548387096774	437\\
325.161290322581	1070\\
336.774193548387	1321\\
348.387096774194	559\\
360	1686\\
};
\end{axis}
\end{tikzpicture}%
}  
  \label{fig:hela-hog}}
     \subfloat[][CDT]{   %
     \scalebox{0.3}{
%
%
\definecolor{mycolor1}{rgb}{0.00000,0.44700,0.74100}%
\definecolor{mycolor2}{rgb}{0.85000,0.32500,0.09800}
\begin{tikzpicture}

\begin{axis}[%
width=3.1in,
height=2.479167in,
at={(0.172222in,3.0625in)},
scale only axis,
separate axis lines,
every outer x axis line/.append style={black},
every x tick label/.append style={font=\color{black}},
xmin=0,
xmax=1,
every outer y axis line/.append style={black},
every y tick label/.append style={font=\color{black}},
ymin=-0.2,
ymax=0.05
]
\addplot [color=mycolor1,solid,line width=2.0pt,forget plot]
  table[row sep=crcr]{%
0	0.00435098231114155\\
0.032258064516129	-0.00796191999965865\\
0.0645161290322581	-0.0216207716578582\\
0.0967741935483871	-0.0389591489111285\\
0.129032258064516	-0.0582001461377405\\
0.161290322580645	-0.0771186982764266\\
0.193548387096774	-0.0906973333158227\\
0.225806451612903	-0.103789486515086\\
0.258064516129032	-0.115682154739096\\
0.290322580645161	-0.122121673833087\\
0.32258064516129	-0.128634639733105\\
0.354838709677419	-0.137007699994576\\
0.387096774193548	-0.142370146115222\\
0.419354838709677	-0.149944729245133\\
0.451612903225806	-0.155696067016166\\
0.483870967741935	-0.161340284011559\\
0.516129032258065	-0.168292778315224\\
0.548387096774194	-0.173844064937986\\
0.580645161290323	-0.162575165899729\\
0.612903225806452	-0.160669423823566\\
0.645161290322581	-0.167129360740618\\
0.67741935483871	-0.141955058377566\\
0.709677419354839	-0.103021009688515\\
0.741935483870968	-0.0751194348763697\\
0.774193548387097	-0.0658689622722881\\
0.806451612903226	-0.0552781683709501\\
0.838709677419355	-0.0426997198869959\\
0.870967741935484	-0.00623802884009117\\
0.903225806451613	0.0114821669536435\\
0.935483870967742	0.0102651505831594\\
0.967741935483871	0.0239439130826871\\
1	0.0156250000000002\\
};
\end{axis}

\begin{axis}[%
width=3.1in,
height=2.479167in,
at={(0.172222in,0.291667in)},
scale only axis,
separate axis lines,
every outer x axis line/.append style={black},
every x tick label/.append style={font=\color{black}},
xmin=0,
xmax=1,
every outer y axis line/.append style={black},
every y tick label/.append style={font=\color{black}},
ymin=-0.15,
ymax=0.05
]
\addplot [color=mycolor2,solid,line width=2.0pt,forget plot]
  table[row sep=crcr]{%
0	0.0127220738070754\\
0.032258064516129	0.0153315480103025\\
0.0645161290322581	0.015795341757379\\
0.0967741935483871	0.01808655864408\\
0.129032258064516	0.0161922258429961\\
0.161290322580645	0.00975118232690902\\
0.193548387096774	0.000352274848721634\\
0.225806451612903	-0.0102683197938659\\
0.258064516129032	-0.0217094656115696\\
0.290322580645161	-0.0324940691911246\\
0.32258064516129	-0.043011273798342\\
0.354838709677419	-0.0536311816241694\\
0.387096774193548	-0.0669446665945839\\
0.419354838709677	-0.0815698434879031\\
0.451612903225806	-0.0945653692522287\\
0.483870967741935	-0.106007976272316\\
0.516129032258065	-0.109134693490904\\
0.548387096774194	-0.106642046309978\\
0.580645161290323	-0.110797040578641\\
0.612903225806452	-0.0897068243766876\\
0.645161290322581	-0.0450255609361458\\
0.67741935483871	-0.0197526988495943\\
0.709677419354839	-0.00989737021852755\\
0.741935483870968	-0.00687858402883623\\
0.774193548387097	-0.00833734313634527\\
0.806451612903226	-0.0107634135072822\\
0.838709677419355	-0.0108242567004619\\
0.870967741935484	-0.0014099215535629\\
0.903225806451613	-0.000514122655810301\\
0.935483870967742	-0.00320172109690775\\
0.967741935483871	0.00739705038813609\\
1	0.015625\\
};
\end{axis}
\end{tikzpicture}%
}  
  \label{fig:hela-cdt}}
     \caption{Two classes of HeLa dataset, Actin (top row) vs. Microtubules (bottom row)}
     \label{fig:hela}
\end{figure}

\begin{figure}[h!]
    \centering
    \subfloat[][PLDA projection in $L^2$ space]{
     \scalebox{0.3}{
%
%
\definecolor{mycolor1}{rgb}{0.00000,0.44700,0.74100}%
\definecolor{mycolor2}{rgb}{0.85000,0.32500,0.09800}%
\newcommand\mtlarge{\fontsize{20pt}{24pt}\selectfont}
\begin{tikzpicture}

\begin{axis}[%
width=6.027778in,
height=4.754167in,
at={(1.011111in,0.641667in)},
scale only axis,
every outer x axis line/.append style={black},
every x tick label/.append style={font=\color{black}},
xmin=-600,
xmax=400,
xticklabel style={
        /pgf/number format/fixed,
        /pgf/number format/precision=2
},
max space between ticks = 50,
try min ticks = 8,
every outer y axis line/.append style={black},
every y tick label/.append style={font=\color{black}},
ymin=-9000,
ymax=1000,
axis x line*=bottom,
axis y line*=left,
legend style={legend cell align=left,align=left,draw=black},
label style={font=\mtlarge},   
xlabel={$1^{st}$ Discriminant Direction},
ylabel={$2^{nd}$ Discriminant Direction}, 
ticks=none
]
\addplot[only marks,mark=halfdiamond*,mark options={},mark size=6.5000pt,color=mycolor1] plot table[row sep=crcr,]{%
28.9870693953985	-3091.57929570554\\
-345.435244598663	-3598.99837650802\\
-113.39728424186	-4421.04718392896\\
-177.958477766942	-3042.50979752946\\
-138.471048490156	-4495.48213981183\\
-101.6473344852	-3796.52838041291\\
-53.4913250434831	-2297.40408950208\\
-209.539423262411	-1665.69404566702\\
-190.442140169692	-2518.68213077077\\
-198.604244033248	437.26512662089\\
-139.456636462263	-1320.11723545982\\
-139.550402113268	-3604.99086523535\\
-200.272161356424	-4197.82023305372\\
-124.245801120315	-1982.12479814348\\
-214.879284152084	-8826.21793451483\\
-154.887646660914	-3215.41272106894\\
-505.610441402006	-4861.0020702829\\
-208.084464750335	-3013.82900082768\\
-428.619856450613	-3053.37889003226\\
-311.709053299838	-4345.27854473889\\
};
\addlegendentry{actine};

\addplot[only marks,mark=10-pointed star,mark options={},mark size=6.5000pt,color=mycolor2] plot table[row sep=crcr,]{%
238.484535539008	-5921.81207752465\\
161.269242774553	-1953.47774467823\\
152.096809073299	-3688.79914472849\\
145.824256849464	-2270.6995304725\\
211.258869173326	-5899.87351509912\\
113.61683238132	-4516.88334123293\\
141.142989972807	-2420.64090906324\\
149.437259071226	-2326.1179309651\\
97.8813805690599	-3973.67152598756\\
116.126119343352	-3327.70224739776\\
242.26198055897	-4498.7060457067\\
163.895691033454	-7178.72589766053\\
347.294183261641	-3116.33875069782\\
193.412056510925	-727.055060993661\\
370.304761235554	-5108.07914868156\\
35.5630987450996	-4934.99281466692\\
144.239619428882	-1254.74066649274\\
254.651793615501	-3174.23739775867\\
};
\addlegendentry{microtubles};

\end{axis}
\end{tikzpicture}%
}  
 \label{fig:hela-plda-l2}
            }
    \subfloat[][PLDA projection in CDT space]{ 
     \scalebox{0.3}{
%
%
\definecolor{mycolor1}{rgb}{0.00000,0.44700,0.74100}%
\definecolor{mycolor2}{rgb}{0.85000,0.32500,0.09800}%
\newcommand\mtlarge{\fontsize{20pt}{24pt}\selectfont}
\begin{tikzpicture}

\begin{axis}[%
width=6.027778in,
height=4.754167in,
at={(1.011111in,0.641667in)},
scale only axis,
every outer x axis line/.append style={black},
every x tick label/.append style={font=\color{black}},
xmin=-0.28,
xmax=-0.14,
xticklabel style={
        /pgf/number format/fixed,
        /pgf/number format/precision=2
},
max space between ticks = 50,
try min ticks = 8,
every outer y axis line/.append style={black},
every y tick label/.append style={font=\color{black}},
ymin=-0.3,
ymax=0.2,
axis x line*=bottom,
axis y line*=left,
legend style={legend cell align=left,align=left,draw=black},
label style={font=\mtlarge},   
xlabel={$1^{st}$ Discriminant Direction},
ylabel={$2^{nd}$ Discriminant Direction}, 
ticks=none
]
\addplot[only marks,mark=halfdiamond*,mark options={},mark size=6.5000pt,color=mycolor1] plot table[row sep=crcr,]{%
-0.234060204531196	-0.197573515656804\\
-0.250382778841216	-0.227418833217413\\
-0.274284254354277	0.0766902551156572\\
-0.251796862207225	-0.0408107542416546\\
-0.240712115019155	-0.0564111903897892\\
-0.269091559276105	0.10567755577634\\
-0.256891182565091	0.0855774076836525\\
-0.234219224781091	-0.0209921704129626\\
-0.236398816115235	-0.0998182690251131\\
-0.237143193416323	-0.0963921478047574\\
-0.233859545597135	-0.133334313707357\\
-0.244321680944662	-0.0411209311724505\\
-0.253966935078806	0.0407517073450184\\
-0.252265848875531	-0.0903323240824961\\
-0.251212574720236	0.15887214596152\\
-0.275131148293305	0.0295675739353546\\
-0.272571715045799	-0.275409052974103\\
-0.241780357644178	-0.147666296259906\\
-0.250861715894495	-0.212827792670345\\
-0.24496369638604	-0.134203130239357\\
};
\addlegendentry{actine};

\addplot[only marks,mark=10-pointed star,mark options={},mark size=6.5000pt,color=mycolor2] plot table[row sep=crcr,]{%
-0.172952335923378	0.0739009115987192\\
-0.190865467343848	0.00655946651596644\\
-0.159702326132724	0.0263928125455129\\
-0.192290678079803	0.0158563373224206\\
-0.182142248295755	0.0130955244469696\\
-0.19611187435307	0.107813479527824\\
-0.169019511715914	-0.130161129835323\\
-0.192767591047614	-0.0669617547140486\\
-0.180596853523622	0.0877851832758489\\
-0.183388957738262	0.0422371350130493\\
-0.185530599578963	-0.0774611198850091\\
-0.171159788343398	0.0885653675966814\\
-0.151662399916033	0.0155563748097397\\
-0.188638833144074	-0.0647269100761414\\
-0.207718139441101	0.137041279100001\\
-0.198266406415229	-0.000338157907850955\\
-0.171860472037676	-0.0202555151668218\\
-0.19560742245689	0.0409425363730506\\
};
\addlegendentry{microtubles};

\end{axis}
\end{tikzpicture}%
}  

             \label{fig:hela-plda-cdt}
             }
              \caption{PLDA projection for HeLa dataset}
  \label{fig:hela-plda}
\end{figure}
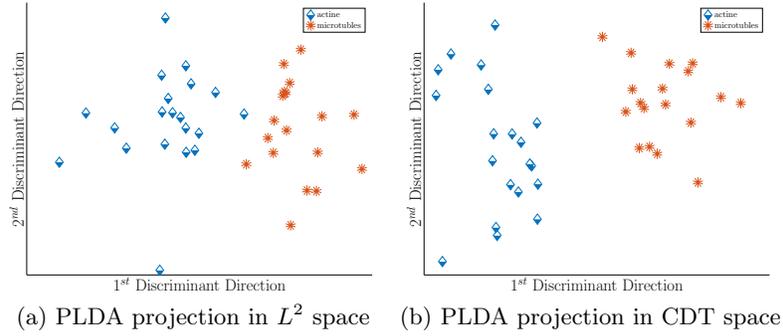

\begin{table}[h!]
\caption{Average Classification Error of the HeLa dataset}
\label{table:heLa}
\centering
\begin{tabular}{|c|c|c|c|}
    \hline
   Classifier type& Dataset & $L^2$ space  &  CDT space\\ 
    \hline
    \multirow{2}{*}{Fisher LDA}       
   & Training set & 0.53 \% & 0.40\% \\
    \hhline{~~~}  
    & Testing set & {\bf 2.66} \% &  {\bf 2.59}\% \\
    \hline
    \multirow{2}{*}{PLDA} 
   & Training set & 0.14 \% & 0.92\% \\
    \hhline{~~~}  
    & Testing set & {\bf 1.59} \% &  {\bf 1.07}\% \\
    \hline
    \multirow{2}{*}{Linear SVM} 
   & Training set & 0 \% & 0.26\% \\
    \hhline{~~~}  
    & Testing set & {\bf 0.53} \% &  {\bf 1.05}\% \\
     \hline
 \end{tabular}
\end{table}

\subsection{Actin and Microtubules Classification}
\label{result:HeLa}
Our goal in this experiment is to quantify how well actin and microtubule filaments in HeLa cells \cite{boland2001neural} differ from one another in terms of their orientation distributions. Fluorescence microscope images of HeLa cells were grouped into two classes according to their protein structure: rhodamine-conjugated phalloidin, which labels F-actin and a monoclonal antibody against beta-tubulin (microtubules). Each image was pre-processed such that outside the cropped region was set to 0 and contrast-stretched to have full scale (see \figref{fig:hela-raw}). In order to compute the orientation of each pixel, the images were filtered with 32 Gabor filters of size $9\times9$, and for each pixel, the filter with the maximum response is selected and labeled from 1 to 32 (see \figref{fig:hela-gabor}). A histogram of orientation filter responses are computed for each image (see \figref{fig:hela-hog}) and then the CDT is computed for each histogram (see \figref{fig:hela-cdt}). In this example, both histogram and CDT show excellent classification accuracy, given that the difference between two protein structures are hard to be recognized by visual inspection. It is an instance where data is already well (linearly) separated in Euclidean space, and is also linearly separable in CDT space (i.e. the CDT did not destroy linear separation in this example).

\section{\textbf{Discussion and conclusions}}
In this paper we have described a new nonlinear operation, termed the Cumulative Distribution Transform (CDT), that takes as input signals that can be understood as probability density functions, and outputs a continuous function that is related to morphing that signal to a chosen reference signal. We have shown that, irrespective of the reference choice, the transform is useful for converting signal variations that are `Lagrangian' in the sense of displacing (transporting) intensities throughout the signal space, to operations that are `Eulerian' in the sense that they become (simpler) adjustments of the intensities, without transporting them, in transform space. This conversion is the basis for our main Theorem \ref{mainLemma} that states the necessary conditions for the CDT to make signal classes linearly separable in transform space. In addition to describing a few of its properties, we have extensively studied the ability of the CDT to improve the linear separability in comparison to the linear separability in original signal space experimentally in five diverse applications involving both simulated and real data. In all examples shown, the results of Theorem \ref{mainLemma} are confirmed. 

The CDT is cheap to compute. Above we described a numerical approximation for discrete signals that is  $O(N)$, with $N$ the length of the signal. Its computational efficiency combined with theoretical and experimental results presented above suggest that the CDT could be a useful tool for building more complex signal pattern recognition systems in a variety of applications. We emphasize that we envision the CDT to be a useful \emph{pre-processing} step in the search for solutions for complex problems. Once data is transformed in CDT space, other techniques (e.g Fourier and Wavelet transforms, Haralick features, etc.) can be applied to the transformed data as well. The fact that the CDT is a mathematically invertible transform ensures that no information will be lost in this step.

The CDT can be compared to the 'feature map' in kernel methods \cite{scholkopf2002learning} considering that the CDT maps the raw data into the CDT space (analogous to the feature space). However, kernel methods avoid explicit formulae of the feature maps, and rather apply kernel function to the raw data. Hence, the data is rarely analyzed in the feature space in kernel methods. On the other hand, CDT has the advantage of directly utilizing the data in the CDT space (feature space). For example, we can invert the linear classifier in the CDT space to that in the raw data space, and by looking at the latter, we can examine which signal characteristic distinguishes two classes of signals. Moreover, kernel methods require the data to be a vector, a discrete sequence. However, for many natural signals such as physiological signals, their nature is continuous. We emphasize that the CDT methods is more continuous friendly than the kernel methods. 

The main limitation of the CDT as a `feature extraction' method is that, as presented, it can only be applied to signals that can be interpreted as probability density functions (hence positive signals). We note, however, that this is not an impediment to its application in a wide variety of data that naturally satisfy the positivity constraint (normalization to a density function can be achieved by a scaling factor). Examples of problems that involve naturally positive data include commodity (e.g. stock) prices \cite{kim2000genetic}, photon counting devices  (e.g. image pixel intensities) \cite{o2012time}, cell counting devices (e.g. flow cytometry) \cite{cheng2007microfluidic, dataset-flowcytometry}, analysis of fMRI signals \cite{logothetis2001neurophysiological}, analysis of frequency densities (e.g. Fourier descriptors) \cite{zhang2001comparative}, orientation filters \cite{freeman1995orientation, dalal2005histograms}, 3D shape or patch-histograms \cite{deselaers2006sparse}, spectral densities \cite{harsanyi1994hyperspectral} and many others. Pattern recognition systems for such applications normally consist of a feature extraction step, and a statistical pattern analysis (e.g. classification) step. In situations where the data being analyzed is naturally positive, the CDT could be used as a step in this pipeline that could simplify (and enhance the performance) subsequent feature extraction and classification. 

Yet another limitation of the CDT model, as stated in Theorem \ref{mainLemma}, is that the linear separability properties depend on the signals being generated from mother signals through the application of a differential, one to one monotonic function with additional restrictions. In certain cases, a physical model for the data can help determine whether the conditions for linear separability in CDT space are applicable. This is the case for the application involving texture discrimination under brightness and contrast variations shown in the introduction. For other applications (e.g. cancer detection from flow cytometry data), however, we have no underlying physical model to determine whether the necessary conditions for linear separability are met. In some cases, the CDT can indeed be a poor match for the problem. One example would be signal/image classification using texton histograms \cite{Varma2005}. The reason being because the independent variable of these signals has no specific order, and thus the meaning of derivatives with respect to the chosen independent variable ordering is not clear. As such, the model we utilize for the signal classes, which depends on the application of a smooth function to a `mother signal', does not quite apply. In such cases, the CDT can still be applied, though we currently offer no information regarding whether the CDT would enhance (or help destroy) linear separability. The variety of examples shown above, however, have helped us confirm the model is applicable, at least to some extent, to not an insignificant number of applications.

Finally, the work presented here is preliminary, and it could be useful to expand it into several directions. One natural direction would be to utilize a similar technique (conversion from `Lagrangian' to `Eulerian' point of view) to simplify pattern recognition for a broader class of signals, including signals that can obtain negative values, as well as 2-dimensional and 3-dimensional images. Yet another direction to follow is to study whether the CDT formalism has any benefit in sampling and signal estimation problems. These, and other extensions, will be the subject of future work.

\section*{Acknowledgements}
Authors acknowledge support from NSF grant CCF 141502. We also wish to thank Dr. Dejan Slepcev, Mathematics, Carnegie Mellon University, for many fruitful discussions.

\section{References}
\bibliographystyle{natbib}
\bibliography{cdt_arxiv_with_changes_v5.bib}

\appendix

\section{Proof for translation property}

\label{appendix:prop:trans}
Consider a probability density ${I}_1: [y_1, y_2] \to \mathbb{R}$, and let ${I}_\mu:  [y_1 +\mu , y_2 +\mu] \to \RR$ represent a translation of the probability density ${I}_1$ by $\mu$, i.e. $I_\mu(x) = I_1(x - \mu)$. 
To find the CDT for $I_\mu$ with respect to the reference probability density $I_0: X \to \RR$, we solve for $f_\mu: X \to  [y_1 + \mu, y_2 + \mu]$:
\begin{align}
\int_{y_1 + \mu}^{f_\mu(x)} I_\mu(\tau)d\tau =  \int_{\inf(X)}^{x} I_0(\tau)d\tau = x.
\label{appendix:eq:trans1}
\end{align}
And similarly, to find the CDT for $I_1$ with respect to the reference $I_0$, we solve for $f_1: X \to  [y_1, y_2]$:
\begin{align}
\int_{y_1}^{f_1(x)} I_1(\tau)d\tau =  \int_{\inf(X)}^{x} I_0(\tau)d\tau = x.
\label{appendix:eq:trans2}
\end{align}
\eqref{appendix:eq:trans1} and \eqref{appendix:eq:trans2}  can be set equal,
\begin{align}
\int_{y_1 + \mu}^{f_\mu(x)} I_\mu(\tau)d\tau  = \int_{y_1}^{f_1(x)} I_1(\tau)d\tau.
\label{appendix:eq:trans3}
\end{align}
By substituting $I_1$ for $I_\mu$ in \eqref{appendix:eq:trans3}, we have
\begin{align}
\int_{y_1 + \mu}^{f_\mu(x)} I_1(\tau - \mu)d\tau  = \int_{y_1}^{f_1(x)} I_1(\tau)d\tau.
\label{appendix:eq:trans4}
\end{align}
By the change of variables theorem, we can substitute $u = \tau-\mu$ in \eqref{appendix:eq:trans4}
\[
\int_{y_1}^{f_\mu(x) - \mu} I_1(u)du = \int_{y_1}^{f_1(x)} I_1(\tau)d\tau.
\]
Since upper limit on left and right side of the integrals are equal, we have $f_\mu(x) = f_1(x) + \mu$. Substituting this into expression for $\widehat{I}_\mu(x) = (f_\mu(x) - x) \sqrt{I_0(x)}$, we have
\begin{align*}
\widehat{I}_\mu(x)= (f_1(x) +\mu - x) \sqrt{I_0(x)}.
\end{align*}
By substituting $\widehat{I}_1(x)=(f_1(x) - x)\sqrt{I_0(x)}$, we have proved the translation property
\[
\widehat{I}_\mu(x)= \widehat{I}_1(x) + \mu\sqrt{I_0(x)}.
\]

\section{Proof for scaling property}
\label{appendix:prop:scaling}
 Consider a probability density ${I}_1: [y_1, y_2] \to \RR$, and let ${I}_a: [y_1/a, y_2/a] \to \RR $ represent a scaling of the probability density ${I}_1$ by $a$, i.e. $I_a(x) = a I_1(a x)$. To find the CDT for $I_a$ with respect to the reference $I_0: X \to \RR$, we solve for $f_a: X \to : [y_1/a, y_2/a] $:
\begin{align}
\int_{y_1/a}^{f_a(x)} I_a(\tau)d\tau = \int_{\inf(X)}^{x} I_0(\tau)d\tau.
\label{appendix:eq:scale1}
\end{align}
And similarly, to find the CDT for $I_1$ with respect to the reference $I_0$, we solve for $f_1: X \to [ y_1 ,  y_2 ] $:
\begin{align}
\int_{y_1}^{f_1(x)} I_1(\tau)d\tau =  \int_{\inf(X)}^{x} I_0(\tau)d\tau
\label{appendix:eq:scale2}
\end{align}
\eqref{appendix:eq:scale1} and \eqref{appendix:eq:scale2} can be set equal,
\begin{align}
\int_{y_1/a}^{f_a(x)} I_a(\tau)d\tau = \int_{y_1}^{f_1(x)} I_1(\tau)d\tau.
\label{appendix:eq:scale3}
\end{align}
By substituting $I_a = a I_1(a x)$ in \eqref{appendix:eq:scale3}, we have 
\begin{align}
\int_{y_1/a}^{f_a(x)} a I_1(a \tau)d\tau = \int_{y_1}^{f_1(x)} I_1(\tau)d\tau.
\label{appendix:eq:scale4}
\end{align}
By the change of variables theorem we can substitute $a \tau =u$, $a d\tau = du$ in \eqref{appendix:eq:scale4},
\begin{equation*}
\int_{y_1}^{a f_a(x)} I_1(u)du =  \int_{y_1}^{f_1(x)} I_1(\tau)d\tau.
\end{equation*}
Since the upper limit on left and right side of the integrals are equal, we have $f_a(x) = \frac{f_1(x)}{a}$. 
Substituting this expression for $\widehat{I}_a (x) = (f_a(x)-x)\sqrt{I_0(x)}$, and cleaning up some algebras, we get $\widehat{I}_a: X \to \RR$:
\[
\widehat{I}_a (x) = \frac{\widehat{I}_1(x) - x(a-1)\sqrt{I_0(x)}}{a}.
\]
 
\section{Proof for composition property}
\label{appendix:prop:comp}
Let $I_1:Y \to \RR$ represent a probability density, and $J_1: Y \to \RR$ its cumulative distribution function. Let $I_g: Z \to \RR$ represent a probability density that has the following relation with $I_1$:
\begin{align}
J_g(x) = J_1(g(x)).
\label{eq:comp}
\end{align}
$J_g: Z \to \RR$ represent the corresponding cumulative distribution for $I_g$, and $g: Z \to Y$ is an invertible, differentiable.
By differentiating each side of \eqref{eq:comp}, we have
 \[
 I_g(x)= g'(x)I_1(g(x)).
 \]
To find the CDT for $I_g$ with respect to the reference probability density $I_0: X \to \RR$, we solve for $f_g: X \to Z$:
\begin{align}
\int_{\inf(Z)}^{f_g(x)} I_g(\tau)d\tau =  \int_{\inf(X)}^{x} I_0(\tau)d\tau
\label{appendix:eq:comp1}
\end{align}
And similarly, to find the CDT for $I_1$, we solve for $f_1: X \to Y$:
\begin{align}
\int_{\inf(Y)}^{f_1(x)} I_1(\tau)d\tau =  \int_{\inf(X)}^{x} I_0(\tau)d\tau
\label{appendix:eq:comp2}
\end{align}
\eqref{appendix:eq:comp1} and \eqref{appendix:eq:comp2} can be set equal,
\begin{align}
\int_{\inf(Z)}^{f_g(x)} I_g(\tau)d\tau   =\int_{\inf(X)}^{f_1(x)} I_1(\tau)d\tau.
\label{appendix:eq:comp3}
\end{align}
By substituting $ I_g(x)= g'(x)I_1(g(x))$ in \eqref{appendix:eq:comp3}, we have 
\begin{align}
\int_{\inf(Z)}^{f_g(x)} g'(\tau) I_1(g( \tau))d\tau  =  \int_{\inf(Y)}^{f_1(x)} I_1(\tau)d\tau.
\label{appendix:eq:comp4}
\end{align}
By the change of variables theorem we can substitute $g (\tau) =u, g'(\tau)d\tau = du $ in \eqref{appendix:eq:comp4},
\begin{equation*}
\int_{\inf(Y)}^{g (f_g(x))} I_1(u)du =  \int_{\inf(Y)}^{f_1(x)} I_1(\tau)d\tau.
\end{equation*}
Since the upper limit on left and right side of the integrals are equal, we have 
\[
g(f_g(x)) = f_1(x).
\]
Since $g$ is an invertible function, $f_g(x) =g^{-1}(f_1(x))$ holds. By substituting this expression for $\widehat{I}_g (x) = (f_g(x)-x)\sqrt{I_0(x)}$, and cleaning up some algebra, we get $\widehat{I}_g: Z \to \RR$:
\begin{align*}
\widehat{I}_g (x) &= \left(g^{-1}\left(f_1(x)\right)-x\right)\sqrt{I_0(x)}\\
 &= \left(g^{-1}\left( \frac{\widehat{I}_1(x)}{\sqrt{I_0(x)}}+x \right)-x\right)\sqrt{I_0(x)}.
\end{align*}

\section{Proof for Lemma \ref{lemma:linearSeparability}}
\label{appendix:linearSeparability}

\begin{proof}
{\it (if)} The convex hulls of compact convex sets are compact in $L^2$ space. Therefore, the convex hulls are compact. For disjoint, compact convex sets sets, we know from Lemma \ref{lemma:linear_classifier} that there exists a hyperplane that linear separates the two. Therefore, if convex hulls are disjoint (i.e. \eqref{convexIneq} holds), then $\mathbb{P}$ and $\mathbb{Q}$ are linearly separable.

{\it (only if)} Suppose $\mathbb{P}$ and $\mathbb{Q}$ are linearly separable but there exists convex hulls of  $\mathbb{P}$ and $\mathbb{Q}$ that are not disjoint, i.e. there exist $ \{p_i \}_{i=1}^{N_p} \subset \mathbb{P}, \{q_j\}_{j=1}^{N_q} \subset \mathbb{Q} $, and $\alpha_i, \beta_j>0$ that satisfies $\sum_{i=1}^{N_p} \alpha_i = 1$, $\sum_{j=1}^{N_q} \beta_j = 1$ s.t.  
\begin{align}
\sum_{i=1}^{N_p} \alpha_i p_i = \sum_{j=1}^{N_q} \beta_j q_j, 
\label{eq:not_disjoint}
\end{align}
for finite $N_p$, $N_q$. We can easily see that this contradicts linear separability. Suppose there exists a linear classifier (i.e. $w(x) =b $ exists that satisfies \eqref{eq:linear_classifier}). By multiplying each side of \eqref{eq:not_disjoint} with $w(x)$ and integrating over $X$, we have 
\begin{align}
\int_X w(x) \left(\sum_i \alpha_i p_i(x)\right)dx   = \int_X w(x) \left(\sum_j \beta_j q_j (x)\right)dx.
\label{eq:not_disjoint_2}
\end{align}
The left side of \eqref{eq:not_disjoint_2} is always smaller than $b$ because 
\begin{align}
\int_X w(x) \left(\sum_i \alpha_i p_i(x)\right)dx = \sum_i \alpha_i  \int_X w(x)p_i(x) dx < \sum_i \alpha_i b =b.
\label{eq:not_disjoint_left}
\end{align}
On the other hand, the right side of \eqref{eq:not_disjoint_2} is always larger than $b$ because 
\begin{align}
\int_X w(x) \left(\sum_j \beta_j q_j (x)\right)dx= \sum_j \beta_j  \int_X w(x)  q_j(x)dx > \sum_j \beta_j b =b
\label{eq:not_disjoint_right}
\end{align}
However, \eqref{eq:not_disjoint_left} and \eqref{eq:not_disjoint_right} contradict to the equivalence in \eqref{eq:not_disjoint_2}, which implies that the linear classifier $w$ cannot exist. Therefore, the convex hulls must be disjoint if linear classifier exists. 
\end{proof}

\section{Proof for Theorem \ref{mainLemma}}
\label{appendix:mainLemma}

\begin{proof} We show that $\widehat{\mathbb{P}}$, $\widehat{\mathbb{Q}}$ must be linearly separable. If not, it would contradict Definition \ref{def:model} that they are disjoint. Suppose $\widehat{\mathbb{P}}$, $\widehat{\mathbb{Q}}$ are not linearly separable. Then by Lemma \ref{lemma:linearSeparability}, there exist $\{p_i\}_{i=1}^{N_p} \subset \mathbb{P}$, $\{q_j\}_{j=1}^{N_q} \subset \mathbb{Q}$, and  $ \alpha_i, \beta_j  > 0$ that satisfies $\;\sum_{i=1}^{N_p}\alpha_i=\sum_{j=1}^{N_q}\beta_j=1$ such that the convex combination of $\{p_i\}_{i=1}^{N_p}$ and $\{q_j\}_{j=1}^{N_q}$ are equivalent, i.e.
\begin{align*}
\sum_{i=1}^{N_p} \alpha_i \widehat{p}_i &= \sum_{j=1}^{N_q} \beta_j \widehat{q}_j.
\end{align*}
By substituting $\widehat{p}_i  = (f_i - \mathbbm{1}) \sqrt{I_0}$ and $\widehat{q}_j  = (g_j - \mathbbm{1}) \sqrt{I_0}$, where $\mathbbm{1}$ refers to an identity map, we have
\begin{align*}
\sum_{i=1}^{N_p} \alpha_i (f_i - \mathbbm{1} )\sqrt{I_0} &= \sum_{j=1}^{N_q} \beta_j (g_j - \mathbbm{1})\sqrt{I_0}.
\end{align*}
By using $\sum_{i=1}^{N_p} \alpha_i = \sum_{j=1}^{N_q} \beta_j = 1$, and dividing each side of the equation by $I_0$, we have
\[
\sum_{i=1}^{N_p} \alpha_i f_i = \sum_{j=1}^{N_q} \beta_j g_j.
\]
By substituting $f_i = h_i^{-1} \circ f_0$ and $g_j = h_j^{-1} \circ g_0$ (see Lemma E.1 presented below), we have 
\begin{align*}
\sum_{i=1}^{N_p} \alpha_i (h_i^{-1} \circ f_0) = \sum_{j=1}^{N_q} \beta_j (h_j^{-1} \circ g_0).
\end{align*}
By substituting $h_{\alpha}^{-1} = \sum_{i=1}^{N_p} \alpha_i h_i^{-1}$ and $h_{\beta}^{-1} = \sum_{j=1}^{N_q} \beta_j h_j^{-1}$, we have
\begin{align*}
h_\alpha^{-1} \circ f_0 = h_\beta^{-1}\circ g_0.
\end{align*}
By composing each side of the equation with $h_\alpha$, we have
\begin{align}
f_0  = h_\alpha \circ h_\beta^{-1} \circ g_0.
\label{eq:f_0}
\end{align}
Note that $h_\alpha^{-1}, h_\alpha, h_\alpha \circ h_\beta^{-1} \in \mathbb{H}$ by conditions {\it i), ii), iii)}.
From the definition of the CDT in \eqref{eq:diffcdt} with respect to reference $I_0$, we have 
\begin{align*}
f_0' (p_0 \circ f_0) &= g_0' (q_0 \circ g_0) = I_0.
\end{align*}
By substituting $f_0$ with the right side of \eqref{eq:f_0}, we have
\begin{alignat*}{2}
&& (h_\alpha \circ h_\beta^{-1} \circ g_0)' (p_0 \circ (h_\alpha \circ h_\beta^{-1} \circ g_0)) & = g_0' (q_0 \circ g_0)\\
\Leftrightarrow && (h_\alpha \circ h_\beta^{-1})' (p_0 \circ (h_\alpha \circ h_\beta^{-1} )) &= q_0\\
\Leftrightarrow && h_{\alpha\beta^{-1}} 'p_0 (h_{\alpha\beta^{-1}}) &= q_0.
\end{alignat*}
The last step of the equation is derive by setting  $h_{\alpha\beta^{-1}} = h_\alpha \circ h_\beta^{-1}$, where $h_{\alpha\beta^{-1}} \in \mathbb{H}$. However, the last statement contradicts the Definition \ref{def:model} that $h'p_0(h)$ and $h'q_0(h)$ each belong to disjoint set $\widehat{\mathbb{P}}$ and $\widehat{\mathbb{Q}}$. Therefore, $\widehat{\mathbb{P}}$, $\widehat{\mathbb{Q}}$ must be linearly separable.

\end{proof}

\begin{lemma2}
Let $f_0$, $f_i$ be \textcolor{black}{monotonic functions} from $X \to Y$ for probability densities $p_0:Y \to \RR$, $p_i:Y \to \RR$ with respect to reference $I_0: X \to \RR$, such that
\begin{align}
\int_{\inf(Y)}^{f_0 (x)} p_0(\tau) d\tau
=\int_{\inf(Y)}^{f_i (x)} p_i(\tau) d\tau=\int_{\inf(X)}^x I_0(\tau)d\tau.
\label{eq:lemma_f}
\end{align}
Then $p_i = h_i'( p_0 \circ h_i)$ implies $h_i \circ f_i = f_0$.
\label{lemma:lemma_f}
\end{lemma2}

\begin{proof}
Substituting \eqref{eq:lemma_f} with $p_i=h'_ip_0(h_i)$, we have
\[
\int_{\inf(Y)}^{f_i(x)} h'_i(\tau)p_0(h_i(\tau))d\tau= \int_{\inf(X)}^x I_0(\tau)d\tau.
\]
By change of variables theorem, substituting $h_i(\tau)=u$ and $h_i'(\tau)d\tau = du$, we have 
\[
 \int_{\inf(Y)}^{h_i(f_i(x))}p_0(u)du =\int_{\inf(X)}^x I_0(\tau)d\tau.
\]
Since $\int_{\inf(Y)}^{f_0(x)}p_0(\tau)d\tau =\int_{\inf(X)}^x I_0(\tau)d\tau$ holds  (see \eqref{eq:lemma_f}), we have
\begin{equation}
\int_{\inf(Y)}^{h_i(f_i(x))}p_0(\tau)d\tau=\int_{\inf(Y)}^{f_0(x)}p_0(\tau)d\tau.
\label{eq:if3}
\end{equation}
The upper limits on each side of the integrals in \eqref{eq:if3} can be set to be equal since both $f_i$ and $h_i$ are strictly increasing functions:
\begin{align}
h_i(f_i(x))=f_0(x).
\label{eq:if4}
\end{align}
Equivalently, we have $f_i(x)=h_i^{-1}(f_0(x))$ by inverting \eqref{eq:if4}.
\end{proof}

\end{document}